\documentclass[11pt]{article}

\usepackage[final]{acl}

\usepackage{microtype}
\usepackage{mathtools}
\usepackage{amsthm}
\usepackage{inconsolata}
\usepackage{tcolorbox}
\usepackage{caption}
\usepackage{adjustbox}
\usepackage{fancyhdr}
\usepackage{url}
\usepackage{empheq, bm, bbm}
\usepackage{booktabs}
\usepackage{multirow}
\usepackage{amsfonts,amsmath,amssymb}

\newcommand{\Loss}[1]{\mathcal{L}\left( #1 \right)}

\newcommand{\NumHypothesis}{K}
\newcommand{\answer}{\texttt{g}}
\newcommand{\answerrepeat}{\answer'}
\newcommand{\answerIndex}{m}
\newcommand{\answerlenght}{M}
\newcommand{\prompt}{\texttt{q}}
\newcommand{\hypothesis}{\texttt{h}}
\newcommand{\token}{w}
\newcommand{\vocab}{\mathcal{V}}
\newcommand{\hypothesisIndex}{k}
\newcommand{\tokenIndex}{i}

\newcommand{\confidence}{c}

\newcommand{\state}{s}
\newcommand{\action}{a}

\newcommand{\policy}[1]{\pi_{#1}}
\newcommand{\BTheta}{\bm{\theta}}

\newcommand{\rewardmodel}{R}

\newcommand{\ARCeasy}{ARC Easy}

\usepackage{amsmath,amsfonts,bm}

\def\eqref#1{equation~\ref{#1}}

\def\1{\bm{1}}

\DeclareMathAlphabet{\mathsfit}{\encodingdefault}{\sfdefault}{m}{sl}
\SetMathAlphabet{\mathsfit}{bold}{\encodingdefault}{\sfdefault}{bx}{n}

\newcommand{\E}{\mathbb{E}}

\DeclareMathOperator*{\argmax}{arg\,max}

\usepackage[english,bidi=default]{babel} 
\babelfont{rm}{TeXGyreTermesX} 
\babelprovide[import]{hindi}
\babelfont[*devanagari]{rm}{Lohit Devanagari}
\babelprovide[import]{arabic}
\babelfont[*arabic]{rm}{Noto Sans Arabic}

\title{Post-Training Large Language Models via\\Reinforcement Learning from Self-Feedback}

\author{Carel van Niekerk, Renato Vukovic, Benjamin Ruppik, \\{\bf Hsien-chin Lin, Shutong Feng, Milica Ga\v{s}i\'{c}} \\
  Heinrich Heine University Düsseldorf, Germany\\
  \texttt{\small{\{revuk100,ruppik,linh,fengs,gasic\}@hhu.de}}\\
  \texttt{\small{academic@carelvniekerk.com}} \\
  }

\usepackage[capitalize,noabbrev]{cleveref}

\theoremstyle{plain}

\theoremstyle{definition}

\theoremstyle{remark}

\usepackage[textsize=tiny]{todonotes}

\begin{document}

\maketitle
\begin{abstract}
  Large Language Models (LLMs) often produce plausible but poorly-calibrated answers, limiting their reliability on reasoning-intensive tasks.
Recent research suggests that Chain-of-Thought (CoT) reasoning paths are inherent in pre-trained LLMs and can be elicited by simply altering the decoding process, where the presence of a CoT path correlates with higher answer confidence. \\
Building on these insights, we present Reinforcement Learning from Self-Feedback (RLSF), a post-training stage that utilises the model’s intrinsic confidence as a self-generated reward.
By generating multiple CoT decoding beams from a frozen LLM, we compute the confidence of each final answer span and rank the resulting traces accordingly to create synthetic preferences.
These preferences are subsequently utilised to fine-tune the policy through standard preference optimisation, requiring no human labels, gold answers, or externally curated rewards. \\
RLSF simultaneously (i) refines the model’s probability estimates--restoring well-behaved calibration--and (ii) strengthens step-by-step reasoning, yielding improved performance on arithmetic reasoning and multiple-choice question answering.
By converting a model’s own uncertainty into structured self-feedback, RLSF affirms reinforcement learning on intrinsic model behaviour as a principled and data-efficient component of the LLM post-training pipeline.
Our results demonstrate that leveraging these inherent reasoning capabilities provides a robust path for enhancing model reliability without manual prompt engineering or external supervision.
\footnote{Code is available under \url{https://github.com/carelvniekerk/ConfidentLLM}.}
\end{abstract}

\section{Introduction}
\label{section:introduction}

\begin{figure}[ht]
    \centering
    \includegraphics[width=0.9\columnwidth]{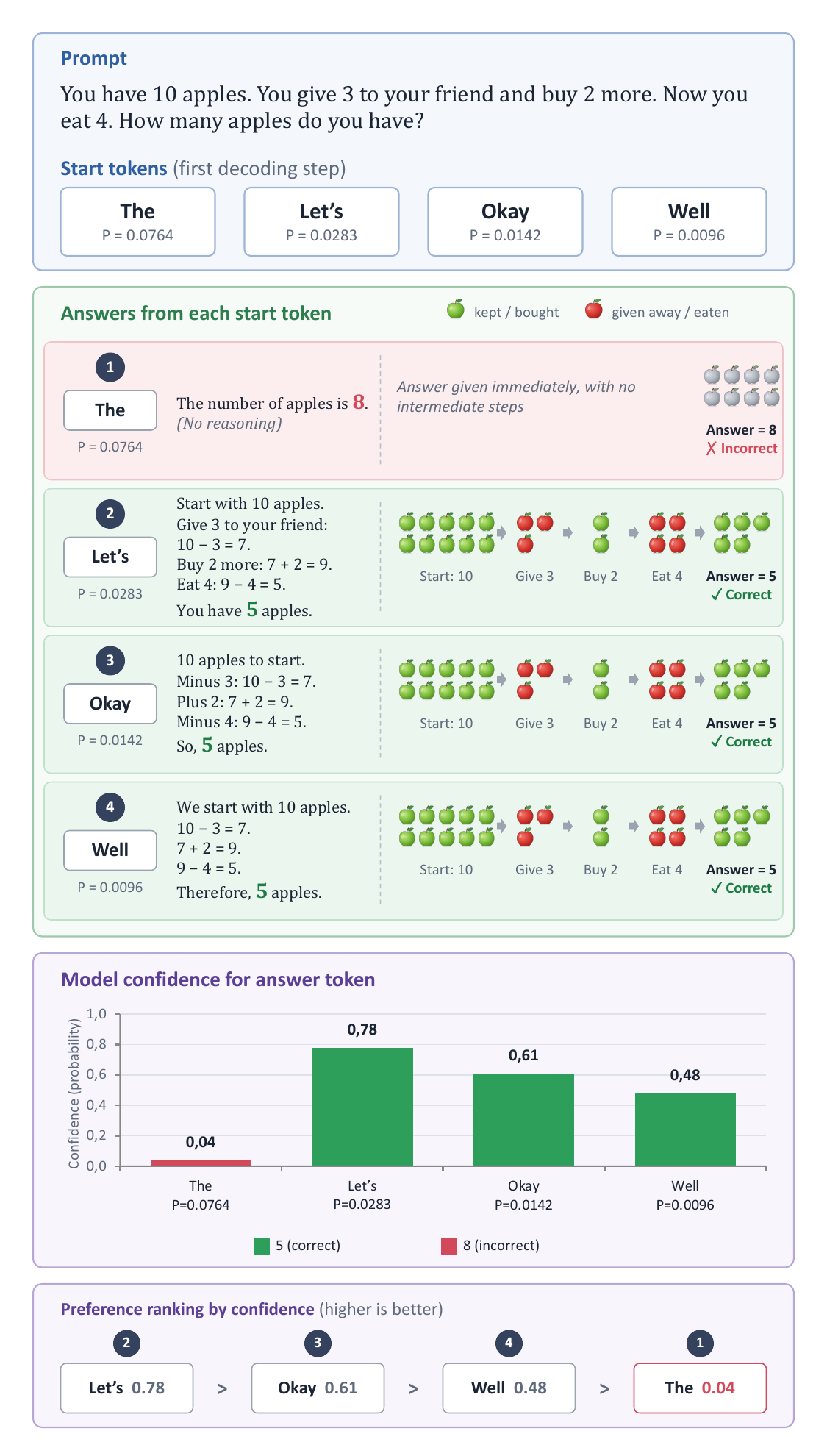}
    \caption{
        An overview of the RLSF preference data generation.
        \label{fig:main-figure}
    }
    \vspace{-10pt}
\end{figure}

Large Language Models (LLMs) have demonstrated remarkable capabilities in text generation and comprehension~\citep{brown2020gpt3,ouyang2022Training}.
However, performance often suffers on tasks requiring extended logical chains, where consistency and accuracy are difficult to maintain during multi-step inference \citep{kambhampati2024CanLarge}.
Furthermore, LLM outputs are frequently poorly calibrated, as expressed confidence rarely predicts actual accuracy~\citep{bai2022Constitutional,tian2023JustAsk,openai2024GPT4Technical}.
This miscalibration leads to overconfidence in complex tasks, making it challenging to identify when a model's output is trustworthy.

Reliable confidence estimates are vital; a model claiming $90\%$ certainty should be correct in $90\%$ of cases.
Yet, standard post-training methods like RLHF, while improving performance, often further degrade calibration~\citep{bai2022Constitutional,tian2023JustAsk,openai2024GPT4Technical}.
This paradox means that common alignment techniques can make models more helpful but simultaneously less reliable.

\paragraph{Inherent Reasoning and Decoding} 

To improve reasoning, Chain-of-Thought (CoT) prompting encourages intermediate steps~\citep{wei2022ChainThought,kojima2022LargeLanguage}, while newer architectures are specifically trained to produce reasoning traces~\citep{openai2024openaio1card}.
Recent research indicates that CoT paths are inherent in pre-trained models and can be elicited simply by altering the decoding process without explicit prompting~\citep{wang2024ChainThoughta}.
Crucially, the presence of these inherent CoT paths correlates with significantly higher confidence in the final answer.
This suggests that token-level confidence is a robust proxy for reasoning quality, effectively separating sound logical paths from flawed sequences.

While CoT-decoding effectively elicits these abilities, the required inference-time compute is an order of magnitude higher than standard methods.
Moreover, most existing calibration techniques rely on labelled correctness data, which severely limits their scalability~\citep{tian2023JustAsk,kapoor2024CalibrationTuning}.

\paragraph{Learning from Inherent Confidence}

We build upon the insight that inherent CoT paths are linked to higher answer confidence.
Instead of using this metric only for inference-time selection, we use it as a signal to generate preference data for model alignment.
By ranking generated reasoning traces based on answer span confidence, we can train a reward model that captures the characteristics of high-quality reasoning.
Similarly, humans can use internal confidence as a feedback mechanism for learning in the absence of external supervision~\citep{ptasczynski2022confidence}.

We introduce \textbf{Reinforcement Learning from Self-Feedback (RLSF)}, a post-training stage using the model's own confidence as an intrinsic reward (\Cref{fig:main-figure}).
RLSF advances self-training in three ways:
(1) it uses the link between decoding confidence and reasoning paths to create synthetic preferences without human labels;
(2) it explicitly optimises for calibration by refining probability estimates via self-generated rewards;
(3) it strengthens step-by-step reasoning through reinforcement learning, achieving the benefits of CoT-decoding at baseline inference costs.

RLSF augments standard post-training pipelines, including supervised fine-tuning and preference optimisation.
While generating CoT beams for training data increases initial costs, the final model maintains a $1\times$ inference speed, making it highly efficient.

\paragraph{Contributions}

We evaluate RLSF on logical reasoning and multiple-choice tasks where accuracy and calibration are typically low.
Our main contributions, confirmed in \Cref{section:experimental_setup,section:results}, are:

\begin{enumerate}
    \item We show that token-level confidence from inherent reasoning paths is an effective reward signal for improving answer-level calibration.
    \item We demonstrate that RLSF provides a practical alternative to CoT-decoding by offering similar performance gains at standard inference costs.
    \item We provide evidence that meaningful reward models can be derived without human labels or external demonstrations using only task-specific prompts.
    \item We confirm that RLSF does not amplify safety-related biases, allowing for safe integration into standard post-training workflows.
\end{enumerate}
\section{Related Work}
\label{section:related_work}

\subsection{Confidence Estimation in LLMs}
\label{subsection:related_work:confidence_based_methods}

Well-calibrated confidence estimates are essential for the reliability of LLMs in decision-support tasks~\citep{wang2024ChainThoughta}.
While supervised fine-tuning often preserves token-level calibration~\citep{kuhn2022Semantic,xiao2022Uncertainty}, alignment via RLHF frequently degrades it, as reward signals prioritise human preference over calibrated uncertainty~\citep{tian2023JustAsk,openai2024GPT4Technical}.
Recent work has moved from single-token metrics to span-level confidence, which better reflects the correctness of complex reasoning chains~\citep{kojima2022LargeLanguage,wang2024ChainThoughta}.
However, existing methods to restore calibration---such as verbalised confidence~\citep{tian2023JustAsk}, auxiliary predictors~\citep{huang2025AccuracyRole}, or calibration-aware tuning~\citep{kapoor2024CalibrationTuning,stangel2025rewarding}---remain limited by their dependence on labelled correctness data.

\subsection{Chain-of-Thought Reasoning}
\label{subsection:related_work:cot_decoding}

Chain-of-Thought (CoT) techniques improve reasoning by prompting models to articulate intermediate steps~\citep{wei2022ChainThought,kojima2022LargeLanguage} or through specific training on reasoning traces~\citep{chung2024Scaling,zelikman2022STaR}.
Notably, \citet{wang2024ChainThoughta} demonstrated that CoT paths are inherent in pre-trained models and can be elicited at inference time by selecting decoding paths with the highest token-level confidence.
\citet{vukovic-etal-2024-dialogue} utilise a constrained version of CoT-decoding for dialogue relation extraction.
While this CoT-decoding bypasses the need for intensive prompt engineering or task-specific fine-tuning, it incurs an order-of-magnitude increase in inference cost, limiting its practical deployment.

\subsection{Preference Learning and Self-Training}
\label{subsection:related_work:preference_learning}

Standard alignment involves reinforcement learning from human feedback~\citep[RLHF;][]{christiano2017Deep, ouyang2022Training} or Direct Preference Optimisation~\citep[DPO;][]{rafailov2023Direct}.
To reduce reliance on human labels, RLAIF~\citep{lee2023RLAIFScaling} and AutoPM~\citep{huang2023Learning} use AI-generated feedback to construct preference datasets.
Self-training methods, such as REST~\citep{gulcehre_reinforced_2023} and AfD~\citep{sun2025InverseRLignment}, further leverage model-generated outputs to improve task accuracy.
RLSF builds on these approaches by using intrinsic confidence as a zero-shot preference signal, eliminating the need for external LLM judges or gold labels.

\subsection{Intrinsic Motivation and Self-Reward}
\label{subsection:related_work:intrinsic_motivation}

Intrinsic rewards allow reinforcement learning to proceed in environments where extrinsic feedback is sparse or absent~\cite{chantanez2004intrinsicallymotivatedRL,barto2013intrinsic}.
In the context of LLMs, this has been explored through self-generated signals where models evaluate their own outputs~\citep{klissarov2024motif}.
Drawing from cognitive science, confidence-based learning is recognised as a fundamental mechanism for human learning without external supervision~\cite{ptasczynski2022confidence}, suggesting a principled basis for using internal confidence as a training signal.

\subsection{Bias and Safety Evaluation}
\label{subsection:related_work:bias_safety}

Evaluating whether post-training methods amplify harmful biases is critical for responsible deployment.
Benchmarks like XSTest~\citep{rottger-etal-2024-xstest} detect exaggerated safety behaviours, while AlpacaEval~\citep{Li_AlpacaEval_An_Automatic_2023} provides scalable approximations to human evaluation using LLM-as-judge frameworks.
Ensuring that RLSF maintains safety profiles while enhancing reasoning is a key component of our evaluation.

\section{Methodology}
\label{section:methodology}

Our method builds on the observation from~\citet{wang2024ChainThoughta} that in well-calibrated models, answer confidence correlates with reasoning quality and accuracy.
We extend this finding by using confidence as an intrinsic reward signal to train models that generate well-calibrated responses.
This is formulated as a sequential decision-making process optimised via reinforcement learning to maximise the confidence of answer tokens.

\paragraph{Notation} 
An auto-regressive large language model can be viewed as a policy \(\policy{\BTheta}\) with parameters \(\BTheta\) in the reinforcement learning framework.
At the initial time step, the state \(\state_0\) is the initial input (prompt or a question) \(\prompt\). 
At each following time step \(t\), given state \(\state_t\) (the prompt and previously generated tokens), the policy selects action \(\action_t\) (the next token \(\token \in \vocab\)) according to the distribution \(\policy{\BTheta}(\action \mid \state)\).
The state transitions to \(\state_{t+1} = \state_t \odot \action_t\), where \(\odot\) denotes concatenation.
Policy parameters are optimised via proximal policy optimisation~\citep[PPO;][]{schulman2017ProximalPolicy}.

\subsection{Chain-of-Thought Decoding}
\label{subsection:methodology:cot_decoding}

Chain-of-Thought (CoT) decoding~\citep{wang2024ChainThoughta} generates multiple reasoning hypotheses for a given input \(\prompt\), then selects the most confident answer.

For input \(\prompt\), CoT decoding samples the \(\NumHypothesis\) top-probability tokens \(\token^1, \dots, \token^\NumHypothesis\) at the first decoding step from \(\policy{\BTheta}(\token|\prompt)\).
Each token \(\token^\hypothesisIndex\) seeds a hypothesis \(\hypothesis^\hypothesisIndex\), which is then extended via greedy auto-regressive decoding:
\[
\hypothesis^\hypothesisIndex \leftarrow \hypothesis^\hypothesisIndex \odot \argmax_\token \policy{\BTheta}(\token|\hypothesis^\hypothesisIndex)
\]
until the end-of-sequence token is generated.

\paragraph{Answer Span Identification}
To identify the answer span \(\answer^\hypothesisIndex\) within hypothesis \(\hypothesis^\hypothesisIndex\), the text ``\texttt{So the answer is}'' is appended to prompt the model to generate answer tokens \(\answerrepeat_\hypothesisIndex\).
The answer span \(\answer^\hypothesisIndex\) is then located in \(\hypothesis^\hypothesisIndex\) via string matching with \(\answerrepeat_\hypothesisIndex\).

\paragraph{Confidence Disparity}
For an answer span \(\answer\) at position \(\answerIndex\) consisting of \(\answerlenght\) tokens, the confidence disparity \(\confidence\) is:
\begin{empheq}{align}
    \label{eq:confidence_disparity}
    \small
    \begin{split}
        \confidence = \frac{1}{\answerlenght} \sum_{\tokenIndex=0}^{\answerlenght-1} \Big[ 
        & \underbrace{\max_\token \policy{\BTheta}(\token | \state_{\answerIndex+\tokenIndex})}_{\text{most likely token probability}} \\
        & - \underbrace{\max_{\token \ne \argmax \policy{\BTheta}} \policy{\BTheta}(\token | \state_{\answerIndex+\tokenIndex})}_{\text{second most likely token probability}}
        \Big]
    \end{split}
\end{empheq}
where \(\state_{\answerIndex+\tokenIndex} = \prompt \odot \hypothesis_{1:\answerIndex+\tokenIndex}\) is the context up to the token being evaluated.

Probability disparity measures the model's certainty by computing the gap between the top two token probabilities at each position in the answer span, then averaging across all answer tokens.
A large disparity (e.g., 0.9 vs 0.05 = 0.85) indicates the model strongly prefers one token; a small disparity (e.g., 0.4 vs 0.35 = 0.05) indicates uncertainty between alternatives.

\textbf{Why disparity rather than raw probability?}
Raw token probability can be misleadingly high due to vocabulary distribution effects—a token with 0.6 probability might seem confident, but if the second-highest is 0.55, the model is actually uncertain.
Disparity captures this uncertainty by considering the full distribution shape.
It has been shown to be more reliable than raw probability~\citep{wang2024ChainThoughta}.

\subsection{Reinforcement Learning from Self-Feedback}
\label{subsection:methodology:rlsf}

Given an input \(\prompt\), CoT decoding produces \(\NumHypothesis\) hypotheses \(\hypothesis^1, \dots, \hypothesis^\NumHypothesis\), each with confidence disparity \(\confidence_1, \dots, \confidence_\NumHypothesis\).
We rank these hypotheses by confidence to construct a preference dataset \(\mathcal{D}\), where higher-confidence responses are treated as preferred over lower-confidence ones.

\paragraph{RLSF with PPO}
We train a reward model \(\rewardmodel_\phi\) on \(\mathcal{D}\) using the Bradley-Terry preference framework~\citep{bradley1952RankAnalysis}, then optimise the policy \(\policy{\BTheta}\) via PPO (see Appendix~\ref{subsection:methodology:ppo} for details).
This approach, which we term RLSF with PPO, follows the standard RLHF pipeline but uses self-generated preferences rather than human annotations.

\paragraph{RLSF with DPO}
Alternatively, \(\mathcal{D}\) can be used to directly fine-tune the model via Direct Preference Optimisation~\citep[DPO;][]{rafailov2023Direct}, eliminating the separate reward model.
We term this RLSF with DPO, though strictly speaking it does not involve reinforcement learning.

\paragraph{Computational Cost}
Training RLSF requires generating \(\NumHypothesis\) CoT beams per prompt to construct \(\mathcal{D}\), incurring K-fold cost during the preference data collection phase.
However, once trained, the RLSF model performs standard greedy decoding at inference time with no additional overhead—unlike CoT-decoding which maintains K-fold cost at inference.

Figure~\ref{fig:main-figure} illustrates the complete RLSF pipeline.

\begin{table*}[t!]
    \centering
    \resizebox{0.95\linewidth}{!}{%
    \begin{tabular}{l l l l l c cc cc}
        \toprule
        \textbf{Base Model} & \textbf{Further} & \textbf{Reward} & \textbf{Prompting} & \textbf{Decoding} & \textbf{Accessible}
                            & \multicolumn{2}{c}{\textbf{MultiArith}} 
                            & \multicolumn{2}{c}{\textbf{GSM8K}} \\
                            & \textbf{optimisation} & \textbf{Model}                      & \textbf{Strategy}                               & \textbf{Strategy}                           & \textbf{Reward Model}
                            & Accuracy \(\uparrow\) & ECE \(\downarrow\)     & Accuracy \(\uparrow\) & ECE \(\downarrow\) \\
        \midrule
        \multirow{6}{*}{\textbf{QWEN 2.5 7B}} 
            & None & DeepSeek & Default & Greedy     & N & \textbf{79.44} & 20.75 & 53.67 & 42.86 \\
            & None & DeepSeek & Default & CoT (10)   & N & 77.77 & 22.22 & \textbf{58.52} & \textbf{41.46} \\
            & None & DeepSeek & CoT     & Greedy     & N & 67.77 & 31.61 & 44.73 & 50.23 \\
            & RLHF(PPO) & URM      & Default & Greedy     & Y & 74.13 & 23.24 & 46.82 & 51.24 \\
            & RLSF(DPO) & Ours       & Default & Greedy     & Y & 71.62 & 25.12 & 50.73 & 38.15 \\
            & RLSF(PPO) & Ours     & Default & Greedy     & Y & 78.06 & \textbf{18.35} & \textbf{58.62} & \textbf{41.92} \\
        \midrule
        \multirow{6}{*}{\textbf{Gemma 2 2B}} 
            & None & Gemma    & Default & Greedy     & N & 98.12 & 7.43  & 85.57 & 12.24 \\
            & None & Gemma    & Default & CoT (10)   & N & \textbf{99.01} & \textbf{4.12}  & \textbf{89.18} & \textbf{10.94} \\
            & None & Gemma    & CoT     & Greedy     & N & 98.08 & 7.18  & 85.61 & 12.83 \\
            & RLHF(PPO) & URM      & Default & Greedy     & Y & 97.83 & 12.73 & 82.43 & 17.83 \\
            & RLSF(DPO)& Ours     & Default & Greedy     & Y & 96.13 & 10.52 & 84.74 & 17.43 \\
            & RLSF(PPO) & Ours     & Default & Greedy     & Y & \textbf{98.83} & 7.81  & 88.14 & 12.54 \\
        \bottomrule
    \end{tabular}}
    \caption{
        Comparison of RLSF variants and baselines on MultiArith and GSM8K. 
        \emph{Accuracy} and \emph{Estimated Calibration Error (ECE)} are reported across models, decoding strategies, and training paradigms. 
        All models are based on the RLHF with PPO variants of the base model.\label{table:rlsf_variant_results}
    }
\end{table*}

\section{Experimental Setup}
\label{section:experimental_setup}

We evaluate RLSF on mathematical reasoning and multiple-choice question answering tasks to assess its impact on both calibration and accuracy.

\subsection{Datasets}
\label{subsection:experimental_setup:datasets}

\paragraph{Mathematical Reasoning}
We use Multi-Arith and GSM8K~\citep{cobbe2021Training}, both containing arithmetic word problems requiring multistep computation and reasoning.
Prompts consist of the question only, with no additional formatting.

\paragraph{Multiple-Choice Question Answering}
We use CommonsenseQA~\citep{talmor2019CommonsenseQA} and ARC Easy~\citep{clark2018ThinkYouHave}, which test common-sense and elementary-level reasoning in a constrained answer format.
Prompts contain the question followed by a list of possible answers.

\paragraph{Reward Model Evaluation}
We evaluate reward models on the RewardBench~\citep{lambert2024rewardbenchevaluatingrewardmodels} mathematical reasoning subset.
Each instance contains two AI-generated responses to the same question, labelled as preferred or rejected by human annotators.
We use only the mathematical reasoning subset for models trained on mathematical tasks.

\subsection{Models}
\label{subsection:experimental_setup:models}

We evaluate RLSF across models with varying calibration quality to assess robustness:

\textbf{Multiple-choice experiments:}
\begin{itemize}
    \item Phi-2~\citep{hughes2023Phi2Surprising} -- a poorly-calibrated model on CommonsenseQA
    \item Gemma 2 2B IT~\citep{team2024Gemma2} -- a well-calibrated model on ARC Easy
\end{itemize}

\textbf{Mathematical reasoning experiments:}
\begin{itemize}
    \item Gemma 2 2B IT~\citep{team2024Gemma2} -- well-calibrated baseline
    \item QWEN 2.5 7B~\citep{deepseek-ai2025DeepSeekR1} -- fine-tuned with DeepSeek R1 distillation
\end{itemize}

Model selection deliberately spans well-calibrated and poorly-calibrated cases to test RLSF's effectiveness under different initial conditions.

\subsection{Metrics}
\label{subsection:experimental_setup:metrics}

\paragraph{Answer Accuracy} 
The percentage of model responses that exactly match correct answers, assessed via numeric equivalence for arithmetic tasks and exact string matching for multiple-choice questions.

\paragraph{Expected Calibration Error (ECE)} 
Measures alignment between confidence and correctness by binning predictions by confidence level and computing the weighted average of absolute differences between confidence and accuracy per bin.
For arithmetic tasks, ECE is computed on answer span tokens; for multiple-choice, on the choice token.
Lower ECE indicates better calibration.

\paragraph{Reward Model Accuracy} 
On RewardBench, the percentage of instance pairs where the reward model correctly assigns a higher score to the human-preferred response.
This evaluates whether RLSF-derived reward models capture meaningful quality signals despite lacking access to ground-truth labels or preferences.

\subsection{Implementation Details}
\label{subsec:experiments:implementation}

\paragraph{Preference Data Construction}
For each training prompt, we generate \(K=10\) candidate responses via CoT decoding since its performance plateaus with this choice of K \citep{wang2024ChainThoughta}.
Candidates are ranked by answer confidence disparity to form preference pairs for training.

\paragraph{Reward Model Training}
Each reward model is initialized from the corresponding base LLM with a scalar regression head attached.
The model is fine-tuned using Low-Rank Adaptation~\citep[LoRA;][]{hu2021LoRALowRank} via the TRL library~\citep{vonwerra2022trl}.
Raw reward values are linearly rescaled to \([-1, 1]\) for stable training dynamics.

\paragraph{Policy Optimization}
For RLSF with PPO, we optimize the base model using the trained reward model following standard RLHF procedures.
For RLSF with DPO, we directly optimize the base model on the preference dataset using the DPO loss.
Detailed hyperparameters are provided in Appendix~\ref{sec:appendix:hyp}.

\paragraph{Baselines}
We compare RLSF against:
\begin{itemize}
    \item \textbf{Base model with greedy decoding} -- Standard inference baseline
    \item \textbf{CoT prompting} -- Prepending ``Let's think step by step'' to prompts
    \item \textbf{CoT decoding (K=10)} -- Generating 10 reasoning paths and selecting the highest-confidence answer~\citep{wang2024ChainThoughta}
    \item \textbf{RLHF with URM} -- Training with the state-of-the-art URM reward model~\citep{lou2025Uncertaintyaware}
\end{itemize}

All experiments use greedy decoding at inference unless otherwise specified.
We do not use sampling-based decoding (temperature, top-p) as our focus is calibration rather than maximum task accuracy, and preliminary experiments showed similar calibration patterns across decoding strategies.

\begin{figure*}[t!]
    \centering
    \includegraphics[width=0.8\linewidth]{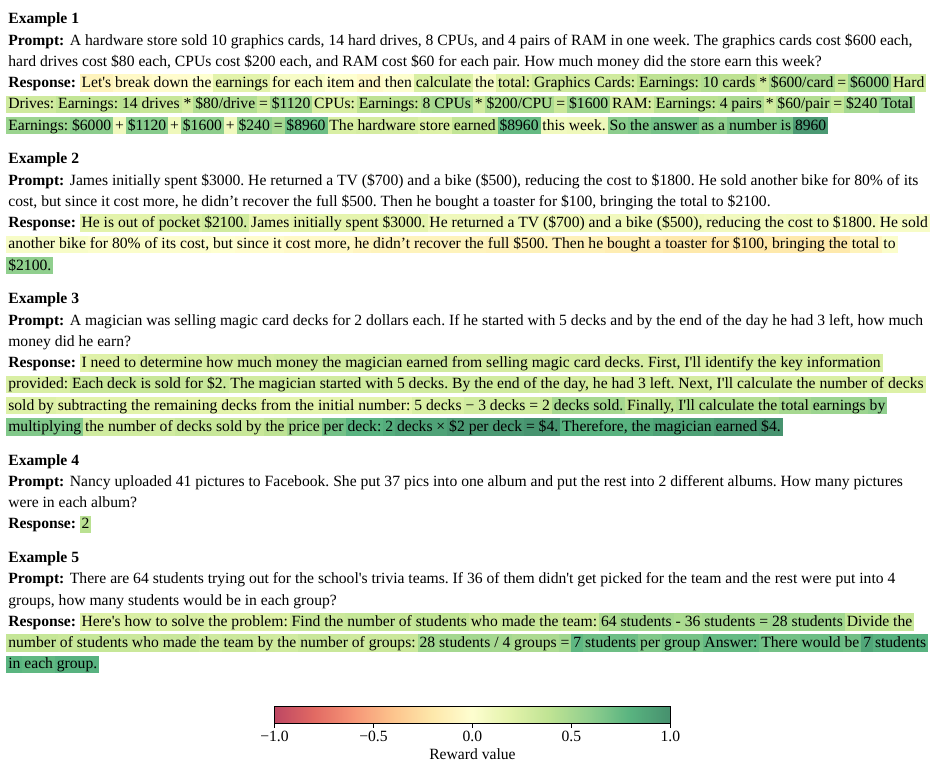}
    \caption{
        Example responses from the RLSF fine-tuned Gemma 2 model. The response words are colored based on the rewards obtained from the RLSF reward model.
        \label{fig:reward_example}
    }
    \vspace{-10pt}
\end{figure*}

\section{Results}
\label{section:results}

We first evaluate RLSF's impact on calibration and accuracy (\Cref{subsection:results_and_discussion:mathematical_reason,subsection:results_and_discussion:multiple_choice}), then validate that RLSF-derived reward models capture meaningful quality signals (\Cref{subsection:results_and_discussion:reward_benchmark}) and finally assess bias amplification (\Cref{subsection:results_and_discussion:bias}).

\subsection{Mathematical Reasoning Tasks}
\label{subsection:results_and_discussion:mathematical_reason}

Table~\ref{table:rlsf_variant_results} compares RLSF against baselines on MultiArith and GSM8K.
We evaluate two base models: QWEN 2.5 7B (fine-tuned with DeepSeek R1 distillation) and Gemma 2 2B (instruction-tuned).
Baselines include CoT prompting, CoT decoding (K=10), and RLHF with URM.

\paragraph{QWEN 2.5 7B (Poorly-Calibrated Base)}
RLSF with PPO achieves the best GSM8K accuracy (58.62\%, matching CoT decoding) and lowest MultiArith ECE (18.35\%).
URM underperforms RLSF on both metrics despite higher RewardBench scores.
CoT prompting performs worst, confirming that inference-time prompting alone is ineffective~\citep{saparov2022LanguageModels}.
RLSF with DPO improves calibration modestly but underperforms the PPO variant.

\paragraph{Gemma 2 2B (Well-Calibrated Base)}
CoT decoding achieves highest accuracy (99.01\% MultiArith, 89.18\% GSM8K) and best calibration (4.12\% ECE MultiArith).
RLSF with PPO ranks second (98.83\% acc, 7.81\% ECE) but eliminates the K-fold inference cost.
This represents a calibration-efficiency trade-off: RLSF trades slight calibration degradation (7.81\% vs 4.12\% ECE) for order-of-magnitude inference speedup.

\paragraph{When Does RLSF Help?}
Results reveal a clear pattern: RLSF provides substantial improvements for poorly-calibrated models (QWEN: 53.67\%→58.62\% acc, 42.86\%→41.92\% ECE on GSM8K) but offers primarily computational benefits for well-calibrated models (Gemma 2).
RLSF is most valuable when: (1) base calibration is degraded post-RLHF, (2) K-fold decoding cost is prohibitive, or (3) calibration is prioritised over maximum accuracy.

\begin{table}[t!]
    \centering
    \resizebox{0.95\linewidth}{!}{%
    \begin{tabular}{l l cc cc}
        \toprule
        \multirow{2}{*}{\textbf{Model}} & \textbf{Decoding} 
                      & \multicolumn{2}{c}{\textbf{Commonsense QA}} 
                      & \multicolumn{2}{c}{\textbf{\ARCeasy}} 
        \\
                      & \textbf{Strategy} & Accuracy \(\uparrow\) & ECE \(\downarrow\) 
                      & Accuracy \(\uparrow\) & ECE \(\downarrow\) 
        \\
        \midrule
        \multirow{2}{*}{Phi-2}         
                      & Greedy     & 54.46 & 25.12 & 85.18 & 27.03 \\
                      & CoT (10)   & 58.91 & 23.11 & 86.10 & 24.83 \\

        \multirow{2}{*}{Gemma 2} 
                      & Greedy     & 79.6 & 19.01 & 96.96 & 16.12 \\
                      & CoT (10)   & 80.03 & 15.49 & 96.28 & 3.03 \\
        \bottomrule
    \end{tabular}}
    \caption{
        Comparison of decoding strategies for Phi-2 and Gemma 2 2B IT across multiple-choice tasks.
        \label{table:base_model_calibration}
    }
\end{table}

\begin{table}[t!]
    \centering
    \resizebox{0.8\columnwidth}{!}{%
    \begin{tabular}{l l cc}
        \toprule
        \textbf{Training} & \textbf{Decoding} 
        & \multirow{2}{*}{Accuracy \(\uparrow\)} & \multirow{2}{*}{ECE \(\downarrow\)} 
        \\
        \textbf{Strategy} & \textbf{Strategy} & & \\
        \midrule
        No           & Greedy     & 96.96 & 16.12 \\
        No           & CoT (10)   & 96.28 & \textbf{3.03} \\
        RLSF (DPO)   & Greedy     & \textbf{97.05} & 18.83 \\
        RLSF (PPO)   & Greedy     & \textbf{97.04} & 5.12 \\
        \bottomrule
    \end{tabular}}
    \caption{
        Performance on {\ARCeasy} using Gemma 2 under different training and decoding strategies.
        \label{table:arc_easy_gemma2}
    }
\end{table}

\begin{table}[t]
    \centering
    \resizebox{0.99\columnwidth}{!}{%
    \begin{tabular}{l cccc}
        \toprule
        \textbf{Reward Model} & 
        \textbf{Prompt} & \textbf{Answer} & \textbf{Preference} & 
        \textbf{Accuracy} \(\uparrow\) \\
        \midrule
        URM LLaMa 3.1 8B     & Y & N & Y & 97.00 \\
        QRM LLaMa 3.1 8B   & Y & N & Y & 96.80 \\
        QWEN 2.5 7B AfD      & Y & Y & N & 89.29 \\
        GEMMA 2 2B AfD       & Y & Y & N & 73.12 \\
        \midrule
        QWEN 2.5 7B RLSF     & Y & N & N & 76.13 \\
        GEMMA 2 2B RLSF      & Y & N & N & 81.43 \\
        \bottomrule
    \end{tabular}
    }
    \caption{
        Reward model accuracy on RewardBench (Math Reasoning). Columns indicate which components (Prompt, Answer, Preference) were used as input to the reward model.
        \label{table:reward_model_math_reasoning}
    }
    \vspace{-10pt}
\end{table}

\paragraph{Reward Model Visualization}
Figure~\ref{fig:reward_example} shows token-level rewards from the RLSF reward model.
Correct answers receive high rewards, with intermediate reasoning steps also rewarded (Examples 1, 5).
Verbose explanations receive lower scores (Example 3), indicating learned preference for conciseness.
The model sometimes exhibits post-hoc rationalization (Example 2), suggesting imperfect reasoning.

Figure~\ref{fig:calibration_example2} demonstrates calibration improvement: both pre- and post-RLSF models answer incorrectly, but RLSF reduces overconfidence (0.993→0.712) and increases probability mass on the correct answer (0.004→0.138).

\begin{figure}[t!]
    \centering
    \begin{tcolorbox}[title=Prompt]
\small
You have to do a lot of thinking while studying a new subject, but it is how you gain what?\\
Select one of the following:\\
A. fatigue\\
B. depression\\
C. best way\\
D. weight\\
E. knowledge
    \end{tcolorbox}

    \begin{tcolorbox}[title=Model Output (Before RLSF)]
\small
\textbf{A. fatigue} Gaining knowledge often requires mental effort and can lead to fatigue, as you're putting in the time and energy to understand and process new information.

\textbf{Predicted: A, Probability: 0.993}

\textbf{Golden Answer: E, Probability: 0.004}
    \end{tcolorbox}

    \begin{tcolorbox}[title=Model Output (After RLSF)]
\small
\textbf{A. fatigue} When studying something new, the mental exertion involved can cause fatigue. While the ultimate goal is to gain knowledge, the immediate effect of extended cognitive effort can be tiredness.

\textbf{Predicted: A, Probability: 0.712}

\textbf{Golden Answer: E, Probability: 0.138}
    \end{tcolorbox}

    \caption{
        Example of calibration improvement via RLSF. 
        Both responses incorrectly answer “A” instead of the correct answer “E”, but the post-RLSF model produces a better explanation with lower confidence.
        \label{fig:calibration_example2}
    }
    \vspace{-10pt}
\end{figure}

\subsection{Multiple-Choice Tasks}
\label{subsection:results_and_discussion:multiple_choice}

Table~\ref{table:base_model_calibration} shows CoT decoding improves calibration across models, confirming prior findings~\citep{wang2024ChainThoughta}.
We select Gemma 2 (well-calibrated on ARC Easy) and Phi-2 (poorly-calibrated on CommonsenseQA) for detailed evaluation.

Results (Tables~\ref{table:arc_easy_gemma2},~\ref{table:commonsense_qa_phi2}) show mathematical reasoning findings: RLSF substantially improves poorly-calibrated models (Phi-2: 54.46\%→61.13\% acc, 25.12\%→19.64\% ECE) while offering modest gains for well-calibrated models (Gemma 2: comparable accuracy, worse calibration than CoT decoding but at 1× inference cost).

\subsection{Reward Model Evaluation}
\label{subsection:results_and_discussion:reward_benchmark}

Table~\ref{table:reward_model_math_reasoning} compares RLSF reward models against methods using human preference labels.
URM~\citep{lou2025Uncertaintyaware} and QRM~\citep{dorka2024Quantile} achieve 97\% accuracy using both prompts and preferences.
AfD~\citep{sun2025InverseRLignment} uses prompts and demonstration answers.
RLSF uses prompts alone, achieving 76--81\% accuracy.

This 16--21 point gap reflects the inherent limitation of self-supervised methods: without ground-truth preferences, RLSF relies entirely on base model confidence.
However, RLSF shows lower variance across base models (76\% vs 81\%) compared to AfD (73\% vs 89\%), suggesting confidence disparity provides a more robust signal than output imitation.
Critically, as shown in Table~\ref{table:rlsf_variant_results}, RewardBench scores do not predict downstream utility: URM (97\% on RewardBench) underperforms RLSF on actual tasks (74.13\% acc vs 78.06\% acc on QWEN MultiArith).

\begin{table}[t!]
    \centering
    \resizebox{0.8\columnwidth}{!}{%
    \begin{tabular}{l l cc}
        \toprule
        \textbf{Training} & \textbf{Decoding} 
        & \multirow{2}{*}{Accuracy \(\uparrow\)} & \multirow{2}{*}{ECE \(\downarrow\)} 
        \\
        \textbf{Strategy} & \textbf{Strategy} & & \\
        \midrule
        No           & Greedy     & 54.46 & 25.12 \\
        No           & CoT (10)   & 58.91 & 23.11 \\
        RLSF (DPO)   & Greedy     & 59.79 & 30.91 \\
        RLSF (PPO)   & Greedy     & \textbf{61.13} & \textbf{19.64} \\
        \bottomrule
    \end{tabular}}
    \caption{
        Performance on Commonsense QA using Phi-2 under various training strategies.
        \label{table:commonsense_qa_phi2}
    }
\end{table}

\begin{table}[t!]
    \centering
    \resizebox{0.9\columnwidth}{!}{%
    \begin{tabular}{l c c}
        \toprule
        \textbf{Training Dataset} 
        & \textbf{Evaluation Dataset} 
        & \textbf{Preference (\%) \(\uparrow\)} \\
        \midrule
        GSM8K + MultiArith & XSTest & 50.73 \\
        CommonsenseQA        & XSTest & 51.82 \\
        XSTest                    & XSTest & 63.24 \\
        \bottomrule
    \end{tabular}}
    \caption{
        AlpacaEval win rates on the XSTest dataset.
        \label{table:alpacaeval}
    }
    \vspace{-15pt}
\end{table}

\subsection{Bias Evaluation}
\label{subsection:results_and_discussion:bias}

Table~\ref{table:alpacaeval} evaluates whether RLSF amplifies biases using XSTest prompts~\citep{rottger-etal-2024-xstest} with GPT-4o as judge~\citep{Li_AlpacaEval_An_Automatic_2023}.
Models trained on task-specific data (GSM8K, CommonsenseQA) show ~50\% preference versus base models, indicating no bias amplification.
When trained directly on XSTest, RLSF reinforces cautious tendencies (63.24\% preference), suggesting safety improvements.

\section{Discussion and Future Work}
\label{section:discussion}

RLSF adopts a self-supervised paradigm similar to recent self-training methods but differs in three key aspects.
First, while REST and self-consistency rely on discrete signals like heuristic filters or majority voting, RLSF utilises token-level probability disparity as a continuous signal for gradient-based calibration optimisation.
Second, RLSF explicitly targets calibration alongside task accuracy; on already well-calibrated models such as Gemma 2, it prioritises inference efficiency over marginal accuracy gains.
Third, RLSF uses RL to learn the reasoning process that generates confident answers rather than merely filtering outputs.
This enables calibration improvements even in poorly-calibrated base models, where simply selecting confident outputs would otherwise amplify existing errors.

The primary trade-off of RLSF is the increased training cost incurred by $K$-fold CoT beam generation for preference data construction.
However, once trained, the model performs standard greedy decoding at inference with no overhead, unlike CoT-decoding which maintains $K$-fold costs at deployment.
Consequently, RLSF is particularly valuable for resource-constrained environments that require the reliability of confidence-based decoding without the associated latency.

As a self-supervised step, RLSF augments existing post-training pipelines~\citep{kumar2025llmposttrainingdeepdive} but does not replace RLHF.
It refines inherent behaviours rather than introducing fundamentally new capabilities or correcting deep-seated pre-training biases.
Our evaluation (\Cref{subsection:results_and_discussion:bias}) indicates that RLSF does not amplify safety-related biases, suggesting it can be safely integrated into production workflows.

\paragraph{Future Work}
Future research could explore the iterative application of RLSF, where the model's improved calibration is used to generate even higher-quality preference data in successive rounds.
Additionally, while we focused on arithmetic and multiple-choice tasks, investigating RLSF in more open-ended creative or technical writing domains could reveal how intrinsic confidence scales with subjective quality.
Finally, combining RLSF with extrinsic rewards in a hybrid objective may allow for simultaneous optimisaton of human alignment and mathematical calibration.

\section{Conclusion}
\label{section:conclusion}

We propose Reinforcement Learning from Self-Feedback (RLSF), a post-training method that improves LLM calibration by using token-level confidence disparity as an intrinsic reward signal.
Unlike existing approaches requiring human labels or demonstrations, RLSF derives preferences solely from prompts via Chain-of-Thought decoding.

Experiments on mathematical reasoning (MultiArith, GSM8K) and multiple-choice tasks (CommonsenseQA, ARC Easy) demonstrate that RLSF improves calibration without harming accuracy.
For poorly-calibrated models, RLSF provides substantial gains in both metrics.
For well-calibrated models, RLSF offers a calibration-efficiency trade-off: comparable performance to inference-time aggregation methods but at standard inference cost.

RLSF with PPO outperforms DPO, demonstrating that reinforcement learning effectively incorporates intrinsic motivation for calibration.
The method does not amplify biases when applied to task-specific domains, suggesting safe integration into production pipelines.

\section*{Limitations}

While RLSF demonstrates significant improvements in calibration and reasoning efficiency, several limitations and associated risks remain. 
First, RLSF primarily addresses single-step reasoning trajectories where the model generates a complete chain of thought before reaching a final answer. 
Extending this framework to tasks requiring multi-turn interaction or long-horizon planning remains future work, although intrinsic rewards have shown promise in augmenting long-term RL methods~\citep{zhou2024ArCHerTraining} and goal-oriented dialogue~\citep{wesselman2018curiosity}. 

Second, the construction of preference data relies on the existence of inherent reasoning paths within the base model. 
If a model is severely under-trained or lacks the latent capability to generate coherent reasoning steps, the token-level confidence signal may be too noisy to form a meaningful reward. 
Furthermore, while our experiments span diverse model sizes and initial calibration qualities, our evaluation was conducted on models with up to 7B parameters. 
Validating RLSF on frontier models (exceeding 70B parameters) and even more complex reasoning benchmarks would be necessary to fully confirm its generalisability across the scaling curve. 

Third, RLSF is a self-supervised method that refines a model's internal consistency; it does not introduce external ``ground truth'' knowledge. 
Consequently, if a model is confidently incorrect due to factual gaps in its pre-training data, RLSF may reinforce these errors rather than correcting them. 
This suggests that RLSF is best utilised as a refinement stage for reasoning logic rather than a mechanism for knowledge acquisition.

Finally, relying entirely on intrinsic confidence as a reward introduces specific alignment risks. 
Optimising a policy against its own internal probability distributions can create a closed-loop feedback system that risks over-fitting to the model's idiosyncratic biases or style preferences. 
While our evaluation shows no amplification of safety-related biases in task-specific domains, deploying an unconstrained self-rewarding loop could potentially cause the model to drift from human intent over multiple iterations, generating highly confident but unaligned or sycophantic reasoning paths.

\section*{Acknowledgements}

This work was funded by the European Research Council (ERC) under the Horizon 2020 research and innovation programme (Grant No. STG2018 804636) and by the Ministry of Culture and Science of North Rhine-Westphalia through the Lamarr Fellow Network.

Computational resources were provided by the Centre for Information and Media Technology at Heinrich Heine University Düsseldorf, Google Cloud, and the Noctua 2 high-performance computer at the NHR Center Paderborn Center for Parallel Computing (PC2). The authors gratefully acknowledge the computing time made available on Noctua 2. PC2 is jointly supported by the Federal Ministry of Research, Technology and Space and the state governments participating in the National High-Performance Computing (NHR) joint funding programme.

\bibliography{references}

@inproceedings{vukovic-etal-2024-dialogue,
    title = "{Dialogue Ontology Relation Extraction via Constrained Chain-of-Thought Decoding}",
    author = "Vukovic, Renato  and
      Arps, David  and
      van Niekerk, Carel  and
      Ruppik, Benjamin Matthias  and
      Lin, Hsien-chin  and
      Heck, Michael  and
      Gasic, Milica",
    editor = "Kawahara, Tatsuya  and
      Demberg, Vera  and
      Ultes, Stefan  and
      Inoue, Koji  and
      Mehri, Shikib  and
      Howcroft, David  and
      Komatani, Kazunori",
    booktitle = "Proceedings of the 25th Annual Meeting of the Special Interest Group on Discourse and Dialogue",
    month = sep,
    year = "2024",
    address = "Kyoto, Japan",
    publisher = "Association for Computational Linguistics",
    url = "https://aclanthology.org/2024.sigdial-1.33",
    doi = "10.18653/v1/2024.sigdial-1.33",
    pages = "370--384",
}

@inproceedings{brown2020gpt3,
 author = {Brown, Tom and Mann, Benjamin and Ryder, Nick and Subbiah, Melanie and Kaplan, Jared D and Dhariwal, Prafulla and Neelakantan, Arvind and Shyam, Pranav and Sastry, Girish and Askell, Amanda and Agarwal, Sandhini and Herbert-Voss, Ariel and Krueger, Gretchen and Henighan, Tom and Child, Rewon and Ramesh, Aditya and Ziegler, Daniel and Wu, Jeffrey and Winter, Clemens and Hesse, Chris and Chen, Mark and Sigler, Eric and Litwin, Mateusz and Gray, Scott and Chess, Benjamin and Clark, Jack and Berner, Christopher and McCandlish, Sam and Radford, Alec and Sutskever, Ilya and Amodei, Dario},
 booktitle = {Advances in Neural Information Processing Systems},
 editor = {H. Larochelle and M. Ranzato and R. Hadsell and M.F. Balcan and H. Lin},
 pages = {1877--1901},
 publisher = {Curran Associates, Inc.},
 title = {{Language Models are Few-Shot Learners}},
 url = {https://proceedings.neurips.cc/paper_files/paper/2020/file/1457c0d6bfcb4967418bfb8ac142f64a-Paper.pdf},
 volume = {33},
 year = {2020}
}

@misc{vonwerra2022trl,
  author = {Leandro von Werra and Younes Belkada and Lewis Tunstall and Edward Beeching and Tristan Thrush and Nathan Lambert and Shengyi Huang and Kashif Rasul and Quentin Gallou{\'e}dec},
  title = {TRL: Transformer Reinforcement Learning},
  year = {2020},
  publisher = {GitHub},
  journal = {GitHub repository},
  howpublished = {\url{https://github.com/huggingface/trl}}
}

@inproceedings{stangel2025rewarding,
  author    = {Bani-Harouni, David and Pellegrini, Chantal and Stangel, Paul and {\"O}zsoy, Ege and Zaripova, Kamilia Pinatovna and Navab, Nassir and Keicher, Matthias},
  title     = {{Rewarding Doubt: A Reinforcement Learning Approach to Calibrated Confidence Expression of Large Language Models}},
  booktitle = {International Conference on Learning Representations (ICLR)},
  year      = {2026},
  url       = {https://openreview.net/forum?id=yResLmrVO1}
}

@inproceedings{sutton1999policygradient,
author = {Sutton, Richard S. and McAllester, David and Singh, Satinder and Mansour, Yishay},
title = {Policy gradient methods for reinforcement learning with function approximation},
year = {1999},
publisher = {MIT Press},
address = {Cambridge, MA, USA},
booktitle = {Proceedings of the 13th International Conference on Neural Information Processing Systems},
pages = {1057--1063},
numpages = {7},
location = {Denver, CO},
series = {NIPS'99}
}

@misc{team2024Gemma2,
    title = {Gemma 2: {Improving} {Open} {Language} {Models} at a {Practical} {Size}},
    shorttitle = {Gemma 2},
    url = {http://arxiv.org/abs/2408.00118},
    doi = {10.48550/arXiv.2408.00118},
    urldate = {2025-05-09},
    publisher = {arXiv},
    author = {Riviere, Morgane and Pathak, Shreya and Sessa, Pier Giuseppe and Hardin, Cassidy and Bhupatiraju, Surya and Hussenot, L{\'e}onard and Mesnard, Thomas and Shahriari, Bobak and Ram{\'e}, Alexandre and Ferret, Johan and Liu, Peter and Tafti, Pouya and Friesen, Abe and Casbon, Michelle and Ramos, Sabela and Kumar, Ravin and Lan, Charline Le and Jerome, Sammy and Tsitsulin, Anton and Vieillard, Nino and Stanczyk, Piotr and Girgin, Sertan and Momchev, Nikola and Hoffman, Matt and Thakoor, Shantanu and Grill, Jean-Bastien and Neyshabur, Behnam and Bachem, Olivier and Walton, Alanna and Severyn, Aliaksei and Parrish, Alicia and Ahmad, Aliya and Hutchison, Allen and Abdagic, Alvin and Carl, Amanda and Shen, Amy and Brock, Andy and Coenen, Andy and Laforge, Anthony and Paterson, Antonia and Bastian, Ben and Piot, Bilal and Wu, Bo and Royal, Brandon and Chen, Charlie and Kumar, Chintu and Perry, Chris and Welty, Chris and Choquette-Choo, Christopher A. and Sinopalnikov, Danila and Weinberger, David and Vijaykumar, Dimple and Rogozi{\'n}ska, Dominika and Herbison, Dustin and Bandy, Elisa and Wang, Emma and Noland, Eric and Moreira, Erica and Senter, Evan and Eltyshev, Evgenii and Visin, Francesco and Rasskin, Gabriel and Wei, Gary and Cameron, Glenn and Martins, Gus and Hashemi, Hadi and Klimczak-Pluci{\'n}ska, Hanna and Batra, Harleen and Dhand, Harsh and Nardini, Ivan and Mein, Jacinda and Zhou, Jack and Svensson, James and Stanway, Jeff and Chan, Jetha and Zhou, Jin Peng and Carrasqueira, Joana and Iljazi, Joana and Becker, Jocelyn and Fernandez, Joe and Amersfoort, Joost van and Gordon, Josh and Lipschultz, Josh and Newlan, Josh and Ji, Ju-yeong and Mohamed, Kareem and Badola, Kartikeya and Black, Kat and Millican, Katie and McDonell, Keelin and Nguyen, Kelvin and Sodhia, Kiranbir and Greene, Kish and Sjoesund, Lars Lowe and Usui, Lauren and Sifre, Laurent and Heuermann, Lena and Lago, Leticia and McNealus, Lilly and Soares, Livio Baldini and Kilpatrick, Logan and Dixon, Lucas and Martins, Luciano and Reid, Machel and Singh, Manvinder and Iverson, Mark and G{\"o}rner, Martin and Velloso, Mat and Wirth, Mateo and Davidow, Matt and Miller, Matt and Rahtz, Matthew and Watson, Matthew and Risdal, Meg and Kazemi, Mehran and Moynihan, Michael and Zhang, Ming and Kahng, Minsuk and Park, Minwoo and Rahman, Mofi and Khatwani, Mohit and Dao, Natalie and Bardoliwalla, Nenshad and Devanathan, Nesh and Dumai, Neta and Chauhan, Nilay and Wahltinez, Oscar and Botarda, Pankil and Barnes, Parker and Barham, Paul and Michel, Paul and Jin, Pengchong and Georgiev, Petko and Culliton, Phil and Kuppala, Pradeep and Comanescu, Ramona and Merhej, Ramona and Jana, Reena and Rokni, Reza Ardeshir and Agarwal, Rishabh and Mullins, Ryan and Saadat, Samaneh and Carthy, Sara Mc and Cogan, Sarah and Perrin, Sarah and Arnold, S{\'e}bastien M. R. and Krause, Sebastian and Dai, Shengyang and Garg, Shruti and Sheth, Shruti and Ronstrom, Sue and Chan, Susan and Jordan, Timothy and Yu, Ting and Eccles, Tom and Hennigan, Tom and Kocisky, Tomas and Doshi, Tulsee and Jain, Vihan and Yadav, Vikas and Meshram, Vilobh and Dharmadhikari, Vishal and Barkley, Warren and Wei, Wei and Ye, Wenming and Han, Woohyun and Kwon, Woosuk and Xu, Xiang and Shen, Zhe and Gong, Zhitao and Wei, Zichuan and Cotruta, Victor and Kirk, Phoebe and Rao, Anand and Giang, Minh and Peran, Ludovic and Warkentin, Tris and Collins, Eli and Barral, Joelle and Ghahramani, Zoubin and Hadsell, Raia and Sculley, D. and Banks, Jeanine and Dragan, Anca and Petrov, Slav and Vinyals, Oriol and Dean, Jeff and Hassabis, Demis and Kavukcuoglu, Koray and Farabet, Clement and Buchatskaya, Elena and Borgeaud, Sebastian and Fiedel, Noah and Joulin, Armand and Kenealy, Kathleen and Dadashi, Robert and Andreev, Alek},
    month = oct,
    year = {2024},
    note = {arXiv:2408.00118 [cs]},
}

@misc{deepseek-ai2025DeepSeekR1,
    title = {{DeepSeek}-{R1}: {Incentivizing} {Reasoning} {Capability} in {LLMs} via {Reinforcement} {Learning}},
    shorttitle = {{DeepSeek}-{R1}},
    url = {http://arxiv.org/abs/2501.12948},
    doi = {10.48550/arXiv.2501.12948},
    urldate = {2025-05-09},
    publisher = {arXiv},
    author = {DeepSeek-AI and Guo, Daya and Yang, Dejian and Zhang, Haowei and Song, Junxiao and Zhang, Ruoyu and Xu, Runxin and Zhu, Qihao and Ma, Shirong and Wang, Peiyi and Bi, Xiao and Zhang, Xiaokang and Yu, Xingkai and Wu, Yu and Wu, Z. F. and Gou, Zhibin and Shao, Zhihong and Li, Zhuoshu and Gao, Ziyi and Liu, Aixin and Xue, Bing and Wang, Bingxuan and Wu, Bochao and Feng, Bei and Lu, Chengda and Zhao, Chenggang and Deng, Chengqi and Zhang, Chenyu and Ruan, Chong and Dai, Damai and Chen, Deli and Ji, Dongjie and Li, Erhang and Lin, Fangyun and Dai, Fucong and Luo, Fuli and Hao, Guangbo and Chen, Guanting and Li, Guowei and Zhang, H. and Bao, Han and Xu, Hanwei and Wang, Haocheng and Ding, Honghui and Xin, Huajian and Gao, Huazuo and Qu, Hui and Li, Hui and Guo, Jianzhong and Li, Jiashi and Wang, Jiawei and Chen, Jingchang and Yuan, Jingyang and Qiu, Junjie and Li, Junlong and Cai, J. L. and Ni, Jiaqi and Liang, Jian and Chen, Jin and Dong, Kai and Hu, Kai and Gao, Kaige and Guan, Kang and Huang, Kexin and Yu, Kuai and Wang, Lean and Zhang, Lecong and Zhao, Liang and Wang, Litong and Zhang, Liyue and Xu, Lei and Xia, Leyi and Zhang, Mingchuan and Zhang, Minghua and Tang, Minghui and Li, Meng and Wang, Miaojun and Li, Mingming and Tian, Ning and Huang, Panpan and Zhang, Peng and Wang, Qiancheng and Chen, Qinyu and Du, Qiushi and Ge, Ruiqi and Zhang, Ruisong and Pan, Ruizhe and Wang, Runji and Chen, R. J. and Jin, R. L. and Chen, Ruyi and Lu, Shanghao and Zhou, Shangyan and Chen, Shanhuang and Ye, Shengfeng and Wang, Shiyu and Yu, Shuiping and Zhou, Shunfeng and Pan, Shuting and Li, S. S. and Zhou, Shuang and Wu, Shaoqing and Ye, Shengfeng and Yun, Tao and Pei, Tian and Sun, Tianyu and Wang, T. and Zeng, Wangding and Zhao, Wanjia and Liu, Wen and Liang, Wenfeng and Gao, Wenjun and Yu, Wenqin and Zhang, Wentao and Xiao, W. L. and An, Wei and Liu, Xiaodong and Wang, Xiaohan and Chen, Xiaokang and Nie, Xiaotao and Cheng, Xin and Liu, Xin and Xie, Xin and Liu, Xingchao and Yang, Xinyu and Li, Xinyuan and Su, Xuecheng and Lin, Xuheng and Li, X. Q. and Jin, Xiangyue and Shen, Xiaojin and Chen, Xiaosha and Sun, Xiaowen and Wang, Xiaoxiang and Song, Xinnan and Zhou, Xinyi and Wang, Xianzu and Shan, Xinxia and Li, Y. K. and Wang, Y. Q. and Wei, Y. X. and Zhang, Yang and Xu, Yanhong and Li, Yao and Zhao, Yao and Sun, Yaofeng and Wang, Yaohui and Yu, Yi and Zhang, Yichao and Shi, Yifan and Xiong, Yiliang and He, Ying and Piao, Yishi and Wang, Yisong and Tan, Yixuan and Ma, Yiyang and Liu, Yiyuan and Guo, Yongqiang and Ou, Yuan and Wang, Yuduan and Gong, Yue and Zou, Yuheng and He, Yujia and Xiong, Yunfan and Luo, Yuxiang and You, Yuxiang and Liu, Yuxuan and Zhou, Yuyang and Zhu, Y. X. and Xu, Yanhong and Huang, Yanping and Li, Yaohui and Zheng, Yi and Zhu, Yuchen and Ma, Yunxian and Tang, Ying and Zha, Yukun and Yan, Yuting and Ren, Z. Z. and Ren, Zehui and Sha, Zhangli and Fu, Zhe and Xu, Zhean and Xie, Zhenda and Zhang, Zhengyan and Hao, Zhewen and Ma, Zhicheng and Yan, Zhigang and Wu, Zhiyu and Gu, Zihui and Zhu, Zijia and Liu, Zijun and Li, Zilin and Xie, Ziwei and Song, Ziyang and Pan, Zizheng and Huang, Zhen and Xu, Zhipeng and Zhang, Zhongyu and Zhang, Zhen},
    month = jan,
    year = {2025},
    note = {arXiv:2501.12948 [cs]},
}

@inproceedings{talmor2019CommonsenseQA,
    address = {Minneapolis, Minnesota},
    title = {{CommonsenseQA}: {A} {Question} {Answering} {Challenge} {Targeting} {Commonsense} {Knowledge}},
    shorttitle = {{CommonsenseQA}},
    url = {https://aclanthology.org/N19-1421/},
    doi = {10.18653/v1/N19-1421},
    urldate = {2025-05-09},
    booktitle = {Proceedings of the 2019 {Conference} of the {North} {American} {Chapter} of the {Association} for {Computational} {Linguistics}: {Human} {Language} {Technologies}, {Volume} 1 ({Long} and {Short} {Papers})},
    publisher = {Association for Computational Linguistics},
    author = {Talmor, Alon and Herzig, Jonathan and Lourie, Nicholas and Berant, Jonathan},
    editor = {Burstein, Jill and Doran, Christy and Solorio, Thamar},
    month = jun,
    year = {2019},
    pages = {4149--4158},
}

@misc{clark2018ThinkYouHave,
    title = {Think you have {Solved} {Question} {Answering}? {Try} {ARC}, the {AI2} {Reasoning} {Challenge}},
    shorttitle = {Think you have {Solved} {Question} {Answering}?},
    url = {http://arxiv.org/abs/1803.05457},
    doi = {10.48550/arXiv.1803.05457},
    urldate = {2025-05-09},
    publisher = {arXiv},
    author = {Clark, Peter and Cowhey, Isaac and Etzioni, Oren and Khot, Tushar and Sabharwal, Ashish and Schoenick, Carissa and Tafjord, Oyvind},
    month = mar,
    year = {2018},
    note = {arXiv:1803.05457 [cs]},
}

@inproceedings{rafailov2023Direct,
	title = {Direct {Preference} {Optimization}: {Your} {Language} {Model} is {Secretly} a {Reward} {Model}},
	volume = {36},
	shorttitle = {Direct {Preference} {Optimization}},
	url = {https://proceedings.neurips.cc/paper_files/paper/2023/file/a85b405ed65c6477a4fe8302b5e06ce7-Paper-Conference.pdf?utm_source=chatgpt.com},
	language = {en},
	urldate = {2025-05-07},
	booktitle = {Advances in {Neural} {Information} {Processing} {Systems}},
	author = {Rafailov, Rafael and Sharma, Archit and Mitchell, Eric and Manning, Christopher D. and Ermon, Stefano and Finn, Chelsea},
	month = dec,
	year = {2023},
	pages = {53728--53741},
}

@inproceedings{huang2025AccuracyRole,
	title = {Beyond {Accuracy}: {The} {Role} of {Calibration} in {Self}-{Improving} {Large} {Language} {Models}},
	shorttitle = {Beyond {Accuracy}},
	url = {https://ieeexplore.ieee.org/stamp/stamp.jsp?arnumber=11400862},
	doi = {10.48550/arXiv.2504.02902},
	urldate = {2025-05-07},
	publisher = {IEEE},
	author = {Huang, Liangjie and Li, Dawei and Liu, Huan and Cheng, Lu},
	month = apr,
    booktitle = {IEEE International Conference on Big Data (Big Data)},
	year = {2025},
}

@inproceedings{kapoor2024CalibrationTuning,
	address = {St Julians, Malta},
	title = {Calibration-{Tuning}: {Teaching} {Large} {Language} {Models} to {Know} {What} {They} {Don}`t {Know}},
	shorttitle = {Calibration-{Tuning}},
	url = {https://aclanthology.org/2024.uncertainlp-1.1/},
	urldate = {2025-05-07},
	booktitle = {Proceedings of the 1st {Workshop} on {Uncertainty}-{Aware} {NLP} ({UncertaiNLP} 2024)},
	publisher = {Association for Computational Linguistics},
	author = {Kapoor, Sanyam and Gruver, Nate and Roberts, Manley and Pal, Arka and Dooley, Samuel and Goldblum, Micah and Wilson, Andrew},
	editor = {V{\'a}zquez, Ra{\'u}l and Celikkanat, Hande and Ulmer, Dennis and Tiedemann, J{\"o}rg and Swayamdipta, Swabha and Aziz, Wilker and Plank, Barbara and Baan, Joris and de Marneffe, Marie-Catherine},
	month = mar,
	year = {2024},
	pages = {1--14},
}

@misc{sun2025InverseRLignment,
	title = {Inverse-{RLignment}: {Large} {Language} {Model} {Alignment} from {Demonstrations} through {Inverse} {Reinforcement} {Learning}},
	shorttitle = {Inverse-{RLignment}},
	url = {http://arxiv.org/abs/2405.15624},
	doi = {10.48550/arXiv.2405.15624},
	urldate = {2025-05-07},
	publisher = {arXiv},
	author = {Sun, Hao and van der Schaar, Mihaela},
	month = jan,
	year = {2025},
	note = {arXiv:2405.15624 [cs]},
}

@inproceedings{lee2023RLAIFScaling,
author = {Lee, Harrison and Phatale, Samrat and Mansoor, Hassan and Mesnard, Thomas and Ferret, Johan and Lu, Kellie and Bishop, Colton and Hall, Ethan and Carbune, Victor and Rastogi, Abhinav and Prakash, Sushant},
title = {{RLAIF vs. RLHF: scaling reinforcement learning from human feedback with AI feedback}},
year = {2024},
publisher = {JMLR.org},
booktitle = {Proceedings of the 41st International Conference on Machine Learning},
articleno = {1071},
numpages = {28},
location = {Vienna, Austria},
series = {ICML'24}
}

@inproceedings{hu2021LoRALowRank,
	title = {{LoRA}: {Low}-{Rank} {Adaptation} of {Large} {Language} {Models}},
	shorttitle = {{LoRA}},
	url = {https://openreview.net/forum?id=nZeVKeeFYf9},
    booktitle = {International Conference on Learning Representations (ICLR)},
	language = {en},
	urldate = {2024-09-17},
	author = {Hu, Edward J. and Shen, Yelong and Wallis, Phillip and Allen-Zhu, Zeyuan and Li, Yuanzhi and Wang, Shean and Wang, Lu and Chen, Weizhu},
	month = oct,
	year = {2022},
}

@misc{hughes2023Phi2Surprising,
  title        = {Phi-2: The surprising power of small language models},
  author       = {Mojan Javaheripi and S{\'e}bastien Bubeck},
  year         = {2023},
  howpublished = {Microsoft Research Blog},
  url          = {https://www.microsoft.com/en-us/research/blog/phi-2-the-surprising-power-of-small-language-models/}
}

@misc{cobbe2021Training,
	title = {Training {Verifiers} to {Solve} {Math} {Word} {Problems}},
	url = {http://arxiv.org/abs/2110.14168},
	doi = {10.48550/arXiv.2110.14168},
	urldate = {2024-09-17},
	publisher = {arXiv},
	author = {Cobbe, Karl and Kosaraju, Vineet and Bavarian, Mohammad and Chen, Mark and Jun, Heewoo and Kaiser, Lukasz and Plappert, Matthias and Tworek, Jerry and Hilton, Jacob and Nakano, Reiichiro and Hesse, Christopher and Schulman, John},
	month = nov,
	year = {2021},
	note = {arXiv:2110.14168 [cs]},
}

@inproceedings{zhou2024ArCHerTraining,
author = {Zhou, Yifei and Zanette, Andrea},
title = {ArCHer: training language model agents via hierarchical multi-turn RL},
year = {2024},
publisher = {JMLR.org},
booktitle = {Proceedings of the 41st International Conference on Machine Learning},
articleno = {2574},
numpages = {32},
location = {Vienna, Austria},
series = {ICML'24}
}

@misc{bai2022Constitutional,
	title = {Constitutional {AI}: {Harmlessness} from {AI} {Feedback}},
	shorttitle = {Constitutional {AI}},
	url = {http://arxiv.org/abs/2212.08073},
	doi = {10.48550/arXiv.2212.08073},
	urldate = {2024-09-16},
	publisher = {arXiv},
	author = {Bai, Yuntao and Kadavath, Saurav and Kundu, Sandipan and Askell, Amanda and Kernion, Jackson and Jones, Andy and Chen, Anna and Goldie, Anna and Mirhoseini, Azalia and McKinnon, Cameron and Chen, Carol and Olsson, Catherine and Olah, Christopher and Hernandez, Danny and Drain, Dawn and Ganguli, Deep and Li, Dustin and Tran-Johnson, Eli and Perez, Ethan and Kerr, Jamie and Mueller, Jared and Ladish, Jeffrey and Landau, Joshua and Ndousse, Kamal and Lukosuite, Kamile and Lovitt, Liane and Sellitto, Michael and Elhage, Nelson and Schiefer, Nicholas and Mercado, Noemi and DasSarma, Nova and Lasenby, Robert and Larson, Robin and Ringer, Sam and Johnston, Scott and Kravec, Shauna and Showk, Sheer El and Fort, Stanislav and Lanham, Tamera and Telleen-Lawton, Timothy and Conerly, Tom and Henighan, Tom and Hume, Tristan and Bowman, Samuel R. and Hatfield-Dodds, Zac and Mann, Ben and Amodei, Dario and Joseph, Nicholas and McCandlish, Sam and Brown, Tom and Kaplan, Jared},
	month = dec,
	year = {2022},
	note = {arXiv:2212.08073 [cs]},
}

@misc{openai2024GPT4Technical,
	title = {{GPT}-4 {Technical} {Report}},
	url = {http://arxiv.org/abs/2303.08774},
	doi = {10.48550/arXiv.2303.08774},
	urldate = {2024-09-16},
	publisher = {arXiv},
	author = {OpenAI and Achiam, Josh and Adler, Steven and Agarwal, Sandhini and Ahmad, Lama and Akkaya, Ilge and Aleman, Florencia Leoni and Almeida, Diogo and Altenschmidt, Janko and Altman, Sam and Anadkat, Shyamal and Avila, Red and Babuschkin, Igor and Balaji, Suchir and Balcom, Valerie and Baltescu, Paul and Bao, Haiming and Bavarian, Mohammad and Belgum, Jeff and Bello, Irwan and Berdine, Jake and Bernadett-Shapiro, Gabriel and Berner, Christopher and Bogdonoff, Lenny and Boiko, Oleg and Boyd, Madelaine and Brakman, Anna-Luisa and Brockman, Greg and Brooks, Tim and Brundage, Miles and Button, Kevin and Cai, Trevor and Campbell, Rosie and Cann, Andrew and Carey, Brittany and Carlson, Chelsea and Carmichael, Rory and Chan, Brooke and Chang, Che and Chantzis, Fotis and Chen, Derek and Chen, Sully and Chen, Ruby and Chen, Jason and Chen, Mark and Chess, Ben and Cho, Chester and Chu, Casey and Chung, Hyung Won and Cummings, Dave and Currier, Jeremiah and Dai, Yunxing and Decareaux, Cory and Degry, Thomas and Deutsch, Noah and Deville, Damien and Dhar, Arka and Dohan, David and Dowling, Steve and Dunning, Sheila and Ecoffet, Adrien and Eleti, Atty and Eloundou, Tyna and Farhi, David and Fedus, Liam and Felix, Niko and Fishman, Sim{\'o}n Posada and Forte, Juston and Fulford, Isabella and Gao, Leo and Georges, Elie and Gibson, Christian and Goel, Vik and Gogineni, Tarun and Goh, Gabriel and Gontijo-Lopes, Rapha and Gordon, Jonathan and Grafstein, Morgan and Gray, Scott and Greene, Ryan and Gross, Joshua and Gu, Shixiang Shane and Guo, Yufei and Hallacy, Chris and Han, Jesse and Harris, Jeff and He, Yuchen and Heaton, Mike and Heidecke, Johannes and Hesse, Chris and Hickey, Alan and Hickey, Wade and Hoeschele, Peter and Houghton, Brandon and Hsu, Kenny and Hu, Shengli and Hu, Xin and Huizinga, Joost and Jain, Shantanu and Jain, Shawn and Jang, Joanne and Jiang, Angela and Jiang, Roger and Jin, Haozhun and Jin, Denny and Jomoto, Shino and Jonn, Billie and Jun, Heewoo and Kaftan, Tomer and Kaiser, {\L}ukasz and Kamali, Ali and Kanitscheider, Ingmar and Keskar, Nitish Shirish and Khan, Tabarak and Kilpatrick, Logan and Kim, Jong Wook and Kim, Christina and Kim, Yongjik and Kirchner, Jan Hendrik and Kiros, Jamie and Knight, Matt and Kokotajlo, Daniel and Kondraciuk, {\L}ukasz and Kondrich, Andrew and Konstantinidis, Aris and Kosic, Kyle and Krueger, Gretchen and Kuo, Vishal and Lampe, Michael and Lan, Ikai and Lee, Teddy and Leike, Jan and Leung, Jade and Levy, Daniel and Li, Chak Ming and Lim, Rachel and Lin, Molly and Lin, Stephanie and Litwin, Mateusz and Lopez, Theresa and Lowe, Ryan and Lue, Patricia and Makanju, Anna and Malfacini, Kim and Manning, Sam and Markov, Todor and Markovski, Yaniv and Martin, Bianca and Mayer, Katie and Mayne, Andrew and McGrew, Bob and McKinney, Scott Mayer and McLeavey, Christine and McMillan, Paul and McNeil, Jake and Medina, David and Mehta, Aalok and Menick, Jacob and Metz, Luke and Mishchenko, Andrey and Mishkin, Pamela and Monaco, Vinnie and Morikawa, Evan and Mossing, Daniel and Mu, Tong and Murati, Mira and Murk, Oleg and M{\'e}ly, David and Nair, Ashvin and Nakano, Reiichiro and Nayak, Rajeev and Neelakantan, Arvind and Ngo, Richard and Noh, Hyeonwoo and Ouyang, Long and O'Keefe, Cullen and Pachocki, Jakub and Paino, Alex and Palermo, Joe and Pantuliano, Ashley and Parascandolo, Giambattista and Parish, Joel and Parparita, Emy and Passos, Alex and Pavlov, Mikhail and Peng, Andrew and Perelman, Adam and Peres, Filipe de Avila Belbute and Petrov, Michael and Pinto, Henrique Ponde de Oliveira and Michael and Pokorny and Pokrass, Michelle and Pong, Vitchyr H. and Powell, Tolly and Power, Alethea and Power, Boris and Proehl, Elizabeth and Puri, Raul and Radford, Alec and Rae, Jack and Ramesh, Aditya and Raymond, Cameron and Real, Francis and Rimbach, Kendra and Ross, Carl and Rotsted, Bob and Roussez, Henri and Ryder, Nick and Saltarelli, Mario and Sanders, Ted and Santurkar, Shibani and Sastry, Girish and Schmidt, Heather and Schnurr, David and Schulman, John and Selsam, Daniel and Sheppard, Kyla and Sherbakov, Toki and Shieh, Jessica and Shoker, Sarah and Shyam, Pranav and Sidor, Szymon and Sigler, Eric and Simens, Maddie and Sitkin, Jordan and Slama, Katarina and Sohl, Ian and Sokolowsky, Benjamin and Song, Yang and Staudacher, Natalie and Such, Felipe Petroski and Summers, Natalie and Sutskever, Ilya and Tang, Jie and Tezak, Nikolas and Thompson, Madeleine B. and Tillet, Phil and Tootoonchian, Amin and Tseng, Elizabeth and Tuggle, Preston and Turley, Nick and Tworek, Jerry and Uribe, Juan Felipe Cer{\'o}n and Vallone, Andrea and Vijayvergiya, Arun and Voss, Chelsea and Wainwright, Carroll and Wang, Justin Jay and Wang, Alvin and Wang, Ben and Ward, Jonathan and Wei, Jason and Weinmann, C. J. and Welihinda, Akila and Welinder, Peter and Weng, Jiayi and Weng, Lilian and Wiethoff, Matt and Willner, Dave and Winter, Clemens and Wolrich, Samuel and Wong, Hannah and Workman, Lauren and Wu, Sherwin and Wu, Jeff and Wu, Michael and Xiao, Kai and Xu, Tao and Yoo, Sarah and Yu, Kevin and Yuan, Qiming and Zaremba, Wojciech and Zellers, Rowan and Zhang, Chong and Zhang, Marvin and Zhao, Shengjia and Zheng, Tianhao and Zhuang, Juntang and Zhuk, William and Zoph, Barret},
	month = mar,
	year = {2024},
	note = {arXiv:2303.08774 [cs]},
}

@inproceedings{tian2023JustAsk,
	address = {Singapore},
	title = {Just {Ask} for {Calibration}: {Strategies} for {Eliciting} {Calibrated} {Confidence} {Scores} from {Language} {Models} {Fine}-{Tuned} with {Human} {Feedback}},
	shorttitle = {Just {Ask} for {Calibration}},
	url = {https://aclanthology.org/2023.emnlp-main.330},
	doi = {10.18653/v1/2023.emnlp-main.330},
	urldate = {2024-09-16},
	booktitle = {Proceedings of the 2023 {Conference} on {Empirical} {Methods} in {Natural} {Language} {Processing}},
	publisher = {Association for Computational Linguistics},
	author = {Tian, Katherine and Mitchell, Eric and Zhou, Allan and Sharma, Archit and Rafailov, Rafael and Yao, Huaxiu and Finn, Chelsea and Manning, Christopher},
	editor = {Bouamor, Houda and Pino, Juan and Bali, Kalika},
	month = dec,
	year = {2023},
	pages = {5433--5442},
}

@inproceedings{xiao2022Uncertainty,
	address = {Abu Dhabi, United Arab Emirates},
	title = {Uncertainty {Quantification} with {Pre}-trained {Language} {Models}: {A} {Large}-{Scale} {Empirical} {Analysis}},
	shorttitle = {Uncertainty {Quantification} with {Pre}-trained {Language} {Models}},
	url = {https://aclanthology.org/2022.findings-emnlp.538},
	doi = {10.18653/v1/2022.findings-emnlp.538},
	urldate = {2024-09-16},
	booktitle = {Findings of the {Association} for {Computational} {Linguistics}: {EMNLP} 2022},
	publisher = {Association for Computational Linguistics},
	author = {Xiao, Yuxin and Liang, Paul Pu and Bhatt, Umang and Neiswanger, Willie and Salakhutdinov, Ruslan and Morency, Louis-Philippe},
	editor = {Goldberg, Yoav and Kozareva, Zornitsa and Zhang, Yue},
	month = dec,
	year = {2022},
	pages = {7273--7284},
}

@inproceedings{
    kuhn2022semantic,
    title={{Semantic Uncertainty: Linguistic Invariances for Uncertainty Estimation in Natural Language Generation}},
    author={Lorenz Kuhn and Yarin Gal and Sebastian Farquhar},
    booktitle={International Conference on Learning Representations (ICLR)},
    year={2023},
    url={https://openreview.net/forum?id=VD-AYtP0dve}
}

@article{bradley1952RankAnalysis,
	title = {Rank {Analysis} of {Incomplete} {Block} {Designs}: {I}. {The} {Method} of {Paired} {Comparisons}},
	volume = {39},
	issn = {0006-3444},
	shorttitle = {Rank {Analysis} of {Incomplete} {Block} {Designs}},
	url = {https://www.jstor.org/stable/2334029},
	doi = {10.2307/2334029},
	number = {3/4},
	urldate = {2024-09-16},
	journal = {Biometrika},
	author = {Bradley, Ralph Allan and Terry, Milton E.},
	year = {1952},
	note = {Publisher: [Oxford University Press, Biometrika Trust]},
	pages = {324--345},
}

@misc{schulman2017ProximalPolicy,
	title = {Proximal {Policy} {Optimization} {Algorithms}},
	url = {http://arxiv.org/abs/1707.06347},
	doi = {10.48550/arXiv.1707.06347},
	urldate = {2024-09-13},
	publisher = {arXiv},
	author = {Schulman, John and Wolski, Filip and Dhariwal, Prafulla and Radford, Alec and Klimov, Oleg},
	month = aug,
	year = {2017},
	note = {arXiv:1707.06347 [cs]},
}

@article{chung2024Scaling,
	title = {Scaling {Instruction}-{Finetuned} {Language} {Models}},
	volume = {25},
	issn = {1533-7928},
	url = {http://jmlr.org/papers/v25/23-0870.html},
	number = {70},
	urldate = {2024-09-12},
	journal = {Journal of Machine Learning Research},
	author = {Chung, Hyung Won and Hou, Le and Longpre, Shayne and Zoph, Barret and Tay, Yi and Fedus, William and Li, Yunxuan and Wang, Xuezhi and Dehghani, Mostafa and Brahma, Siddhartha and Webson, Albert and Gu, Shixiang Shane and Dai, Zhuyun and Suzgun, Mirac and Chen, Xinyun and Chowdhery, Aakanksha and Castro-Ros, Alex and Pellat, Marie and Robinson, Kevin and Valter, Dasha and Narang, Sharan and Mishra, Gaurav and Yu, Adams and Zhao, Vincent and Huang, Yanping and Dai, Andrew and Yu, Hongkun and Petrov, Slav and Chi, Ed H. and Dean, Jeff and Devlin, Jacob and Roberts, Adam and Zhou, Denny and Le, Quoc V. and Wei, Jason},
	year = {2024},
	pages = {1--53},
}

@inproceedings{christiano2017Deep,
	title = {Deep {Reinforcement} {Learning} from {Human} {Preferences}},
	volume = {30},
	url = {https://papers.nips.cc/paper_files/paper/2017/hash/d5e2c0adad503c91f91df240d0cd4e49-Abstract.html},
	urldate = {2024-09-12},
	booktitle = {Advances in {Neural} {Information} {Processing} {Systems}},
	publisher = {Curran Associates, Inc.},
	author = {Christiano, Paul F and Leike, Jan and Brown, Tom and Martic, Miljan and Legg, Shane and Amodei, Dario},
	year = {2017},
}

@inproceedings{ouyang2022Training,
	title = {Training language models to follow instructions with human feedback},
	volume = {35},
	url = {https://proceedings.neurips.cc/paper_files/paper/2022/file/b1efde53be364a73914f58805a001731-Paper-Conference.pdf},
	language = {en},
	urldate = {2024-09-12},
	booktitle = {Advances in {Neural} {Information} {Processing} {Systems}},
	author = {Ouyang, Long and Wu, Jeffrey and Jiang, Xu and Almeida, Diogo and Wainwright, Carroll and Mishkin, Pamela and Zhang, Chong and Agarwal, Sandhini and Slama, Katarina and Ray, Alex and Schulman, John and Hilton, Jacob and Kelton, Fraser and Miller, Luke and Simens, Maddie and Askell, Amanda and Welinder, Peter and Christiano, Paul F. and Leike, Jan and Lowe, Ryan},
	month = dec,
	year = {2022},
	pages = {27730--27744},
}

@article{kambhampati2024CanLarge,
	title = {Can large language models reason and plan?},
	volume = {1534},
	issn = {1749-6632},
	url = {https://onlinelibrary.wiley.com/doi/abs/10.1111/nyas.15125},
	doi = {10.1111/nyas.15125},
	language = {en},
	number = {1},
	urldate = {2024-09-11},
	journal = {Annals of the New York Academy of Sciences},
	author = {Kambhampati, Subbarao},
	year = {2024},
	pages = {15--18},
}

@inproceedings{zelikman2022STaR,
	title = {{STaR}: {Bootstrapping} {Reasoning} {With} {Reasoning}},
	volume = {35},
	shorttitle = {{STaR}},
	url = {https://dl.acm.org/doi/10.5555/3600270.3601396},
	language = {en},
	urldate = {2024-09-11},
	booktitle = {Advances in {Neural} {Information} {Processing} {Systems}},
	author = {Zelikman, Eric and Wu, Yuhuai and Mu, Jesse and Goodman, Noah},
	month = dec,
	year = {2022},
	pages = {15476--15488},
}

@inproceedings{wang2024ChainThoughta,
author = {Wang, Xuezhi and Zhou, Denny},
title = {Chain-of-thought reasoning without prompting},
year = {2024},
isbn = {9798331314385},
publisher = {Curran Associates Inc.},
address = {Red Hook, NY, USA},
booktitle = {Proceedings of the 38th International Conference on Neural Information Processing Systems},
articleno = {2123},
numpages = {27},
location = {Vancouver, BC, Canada},
series = {NIPS '24}
}

@inproceedings{kojima2022LargeLanguage,
	title = {Large {Language} {Models} are {Zero}-{Shot} {Reasoners}},
	volume = {35},
	url = {https://dl.acm.org/doi/10.5555/3600270.3601883},
	language = {en},
	urldate = {2024-09-11},
	booktitle = {Advances in {Neural} {Information} {Processing} {Systems}},
	author = {Kojima, Takeshi and Gu, Shixiang (Shane) and Reid, Machel and Matsuo, Yutaka and Iwasawa, Yusuke},
	month = dec,
	year = {2022},
	pages = {22199--22213},
}

@inproceedings{chantanez2004intrinsicallymotivatedRL,
 author = {Chentanez, Nuttapong and Barto, Andrew and Singh, Satinder},
 booktitle = {Advances in Neural Information Processing Systems},
 editor = {L. Saul and Y. Weiss and L. Bottou},
 pages = {},
 publisher = {MIT Press},
 title = {{Intrinsically Motivated Reinforcement Learning}},
 url = {https://proceedings.neurips.cc/paper_files/paper/2004/file/4be5a36cbaca8ab9d2066debfe4e65c1-Paper.pdf},
 volume = {17},
 year = {2004}
}

@incollection{barto2013intrinsic,
  title     = {Intrinsic motivation and reinforcement learning},
  author    = {Barto, Andrew G.},
  booktitle = {{Intrinsically Motivated Learning in Natural and Artificial Systems}},
  editor    = {Baldassarre, Gianluca and Mirolli, Marco},
  pages     = {17--47},
  year      = {2013},
  publisher = {Springer Science+Business Media}
}

@article{ptasczynski2022confidence,
    doi = {10.1371/journal.pcbi.1010580},
    author = {Ptasczynski, Lena Esther AND Steinecker, Isa AND Sterzer, Philipp AND Guggenmos, Matthias},
    journal = {PLOS Computational Biology},
    publisher = {Public Library of Science},
    title = {The value of confidence: Confidence prediction errors drive value-based learning in the absence of external feedback},
    year = {2022},
    month = {10},
    volume = {18},
    url = {https://doi.org/10.1371/journal.pcbi.1010580},
    pages = {1-25},
    number = {10},

}

@inproceedings{huang2023Learning,
    address = {Singapore},
    title = {Learning {Preference} {Model} for {LLMs} via {Automatic} {Preference} {Data} {Generation}},
    url = {https://aclanthology.org/2023.emnlp-main.570/},
    doi = {10.18653/v1/2023.emnlp-main.570},
    urldate = {2025-05-19},
    booktitle = {Proceedings of the 2023 {Conference} on {Empirical} {Methods} in {Natural} {Language} {Processing}},
    publisher = {Association for Computational Linguistics},
    author = {Huang, Shijia and Zhao, Jianqiao and Li, Yanyang and Wang, Liwei},
    editor = {Bouamor, Houda and Pino, Juan and Bali, Kalika},
    month = dec,
    year = {2023},
    pages = {9187--9199},
}

@inproceedings{
    klissarov2024motif,
    title={{Motif: Intrinsic Motivation from Artificial Intelligence Feedback}},
    author={Martin Klissarov and Pierluca D'Oro and Shagun Sodhani and Roberta Raileanu and Pierre-Luc Bacon and Pascal Vincent and Amy Zhang and Mikael Henaff},
    booktitle={The Twelfth International Conference on Learning Representations},
    year={2024},
    url={https://openreview.net/forum?id=tmBKIecDE9},
}

@inproceedings{saparov2022LanguageModels,
  title={{Language Models Are Greedy Reasoners: A Systematic Formal Analysis of Chain-of-Thought}},
  author={Abulhair Saparov and He He},
  booktitle={The Eleventh International Conference on Learning Representations},
  year={2023},
  url={https://openreview.net/forum?id=qFVVBzXxR2V}
}

@INPROCEEDINGS{wesselman2018curiosity,
  author={Wesselmann, Paula and Wu, Yen-Chen and Gašić, Milica},
  booktitle={ICASSP 2019 - 2019 IEEE International Conference on Acoustics, Speech and Signal Processing (ICASSP)}, 
  title={{Curiosity-driven Reinforcement Learning for Dialogue Management}}, 
  year={2019},
  volume={},
  number={},
  pages={7210-7214},
  doi={10.1109/ICASSP.2019.8683033}}

@ARTICLE{kumar2025llmposttrainingdeepdive,
author={Kumar, Komal and Ashraf, Tajamul and Thawakar, Omkar and Anwer, Rao Muhammad and Cholakkal, Hisham and Shah, Mubarak and Yang, Ming-Hsuan and Torr, Phillip H.S. and Khan, Fahad Shahbaz and Khan, Salman},
journal={ IEEE Transactions on Pattern Analysis \& Machine Intelligence },
title={{ LLM Post-Training: A Deep Dive into Reasoning Large Language Models }},
year={2026},
volume={},
number={01},
ISSN={1939-3539},
pages={1-20},
doi={10.1109/TPAMI.2026.3720414},
url = {https://doi.ieeecomputersociety.org/10.1109/TPAMI.2026.3720414},
publisher={IEEE Computer Society},
address={Los Alamitos, CA, USA},
month=aug}

@misc{openai2024openaio1card,
      title={{OpenAI \emph{o1} System Card}}, 
      author={OpenAI and Aaron Jaech and Adam Kalai and Adam Lerer and Adam Richardson and Ahmed El-Kishky and Aiden Low and Alec Helyar and Aleksander Madry and Alex Beutel and Alex Carney and Alex Iftimie and Alex Karpenko and Alex Tachard Passos and Alexander Neitz and Alexander Prokofiev and Alexander Wei and Allison Tam and Ally Bennett and Ananya Kumar and Andre Saraiva and Andrea Vallone and Andrew Duberstein and Andrew Kondrich and Andrey Mishchenko and Andy Applebaum and Angela Jiang and Ashvin Nair and Barret Zoph and Behrooz Ghorbani and Ben Rossen and Benjamin Sokolowsky and Boaz Barak and Bob McGrew and Borys Minaiev and Botao Hao and Bowen Baker and Brandon Houghton and Brandon McKinzie and Brydon Eastman and Camillo Lugaresi and Cary Bassin and Cary Hudson and Chak Ming Li and Charles de Bourcy and Chelsea Voss and Chen Shen and Chong Zhang and Chris Koch and Chris Orsinger and Christopher Hesse and Claudia Fischer and Clive Chan and Dan Roberts and Daniel Kappler and Daniel Levy and Daniel Selsam and David Dohan and David Farhi and David Mely and David Robinson and Dimitris Tsipras and Doug Li and Dragos Oprica and Eben Freeman and Eddie Zhang and Edmund Wong and Elizabeth Proehl and Enoch Cheung and Eric Mitchell and Eric Wallace and Erik Ritter and Evan Mays and Fan Wang and Felipe Petroski Such and Filippo Raso and Florencia Leoni and Foivos Tsimpourlas and Francis Song and Fred von Lohmann and Freddie Sulit and Geoff Salmon and Giambattista Parascandolo and Gildas Chabot and Grace Zhao and Greg Brockman and Guillaume Leclerc and Hadi Salman and Haiming Bao and Hao Sheng and Hart Andrin and Hessam Bagherinezhad and Hongyu Ren and Hunter Lightman and Hyung Won Chung and Ian Kivlichan and Ian O'Connell and Ian Osband and Ignasi Clavera Gilaberte and Ilge Akkaya and Ilya Kostrikov and Ilya Sutskever and Irina Kofman and Jakub Pachocki and James Lennon and Jason Wei and Jean Harb and Jerry Twore and Jiacheng Feng and Jiahui Yu and Jiayi Weng and Jie Tang and Jieqi Yu and Joaquin Qui{\~n}onero Candela and Joe Palermo and Joel Parish and Johannes Heidecke and John Hallman and John Rizzo and Jonathan Gordon and Jonathan Uesato and Jonathan Ward and Joost Huizinga and Julie Wang and Kai Chen and Kai Xiao and Karan Singhal and Karina Nguyen and Karl Cobbe and Katy Shi and Kayla Wood and Kendra Rimbach and Keren Gu-Lemberg and Kevin Liu and Kevin Lu and Kevin Stone and Kevin Yu and Lama Ahmad and Lauren Yang and Leo Liu and Leon Maksin and Leyton Ho and Liam Fedus and Lilian Weng and Linden Li and Lindsay McCallum and Lindsey Held and Lorenz Kuhn and Lukas Kondraciuk and Lukasz Kaiser and Luke Metz and Madelaine Boyd and Maja Trebacz and Manas Joglekar and Mark Chen and Marko Tintor and Mason Meyer and Matt Jones and Matt Kaufer and Max Schwarzer and Meghan Shah and Mehmet Yatbaz and Melody Y. Guan and Mengyuan Xu and Mengyuan Yan and Mia Glaese and Mianna Chen and Michael Lampe and Michael Malek and Michele Wang and Michelle Fradin and Mike McClay and Mikhail Pavlov and Miles Wang and Mingxuan Wang and Mira Murati and Mo Bavarian and Mostafa Rohaninejad and Nat McAleese and Neil Chowdhury and Neil Chowdhury and Nick Ryder and Nikolas Tezak and Noam Brown and Ofir Nachum and Oleg Boiko and Oleg Murk and Olivia Watkins and Patrick Chao and Paul Ashbourne and Pavel Izmailov and Peter Zhokhov and Rachel Dias and Rahul Arora and Randall Lin and Rapha Gontijo Lopes and Raz Gaon and Reah Miyara and Reimar Leike and Renny Hwang and Rhythm Garg and Robin Brown and Roshan James and Rui Shu and Ryan Cheu and Ryan Greene and Saachi Jain and Sam Altman and Sam Toizer and Sam Toyer and Samuel Miserendino and Sandhini Agarwal and Santiago Hernandez and Sasha Baker and Scott McKinney and Scottie Yan and Shengjia Zhao and Shengli Hu and Shibani Santurkar and Shraman Ray Chaudhuri and Shuyuan Zhang and Siyuan Fu and Spencer Papay and Steph Lin and Suchir Balaji and Suvansh Sanjeev and Szymon Sidor and Tal Broda and Aidan Clark and Tao Wang and Taylor Gordon and Ted Sanders and Tejal Patwardhan and Thibault Sottiaux and Thomas Degry and Thomas Dimson and Tianhao Zheng and Timur Garipov and Tom Stasi and Trapit Bansal and Trevor Creech and Troy Peterson and Tyna Eloundou and Valerie Qi and Vineet Kosaraju and Vinnie Monaco and Vitchyr Pong and Vlad Fomenko and Weiyi Zheng and Wenda Zhou and Wes McCabe and Wojciech Zaremba and Yann Dubois and Yinghai Lu and Yining Chen and Young Cha and Yu Bai and Yuchen He and Yuchen Zhang and Yunyun Wang and Zheng Shao and Zhuohan Li},
      year={2024},
      eprint={2412.16720},
      archivePrefix={arXiv},
      primaryClass={cs.AI},
      url={https://arxiv.org/abs/2412.16720}, 
}

@inproceedings{lambert2024rewardbenchevaluatingrewardmodels,
    title = "{{R}eward{B}ench: Evaluating Reward Models for Language Modeling}",
    author = "Lambert, Nathan  and
      Pyatkin, Valentina  and
      Morrison, Jacob  and
      Miranda, LJ  and
      Lin, Bill Yuchen  and
      Chandu, Khyathi  and
      Dziri, Nouha  and
      Kumar, Sachin  and
      Zick, Tom  and
      Choi, Yejin  and
      Smith, Noah A.  and
      Hajishirzi, Hannaneh",
    editor = "Chiruzzo, Luis  and
      Ritter, Alan  and
      Wang, Lu",
    booktitle = "Findings of the Association for Computational Linguistics: NAACL 2025",
    month = apr,
    year = "2025",
    address = "Albuquerque, New Mexico",
    publisher = "Association for Computational Linguistics",
    url = "https://aclanthology.org/2025.findings-naacl.96/",
    doi = "10.18653/v1/2025.findings-naacl.96",
    pages = "1755--1797",
    ISBN = "979-8-89176-195-7"
}

@inproceedings{rottger-etal-2024-xstest,
    title = "{{XST}est: A Test Suite for Identifying Exaggerated Safety Behaviours in Large Language Models}",
    author = {R{\"o}ttger, Paul  and
      Kirk, Hannah  and
      Vidgen, Bertie  and
      Attanasio, Giuseppe  and
      Bianchi, Federico  and
      Hovy, Dirk},
    editor = "Duh, Kevin  and
      Gomez, Helena  and
      Bethard, Steven",
    booktitle = "Proceedings of the 2024 Conference of the North American Chapter of the Association for Computational Linguistics: Human Language Technologies (Volume 1: Long Papers)",
    month = jun,
    year = "2024",
    address = "Mexico City, Mexico",
    publisher = "Association for Computational Linguistics",
    url = "https://aclanthology.org/2024.naacl-long.301/",
    doi = "10.18653/v1/2024.naacl-long.301",
    pages = "5377--5400"
}

@software{Li_AlpacaEval_An_Automatic_2023,
    author = {Li, Xuechen and Zhang, Tianyi and Dubois, Yann and Taori, Rohan and Gulrajani, Ishaan and Guestrin, Carlos and Liang, Percy and Hashimoto, Tatsunori B.},
    month = may,
    title = {{AlpacaEval: An Automatic Evaluator of Instruction-following Models}},
    year = {2023}
}

@misc{lou2025Uncertaintyaware,
    title = {{Uncertainty-aware {Reward} {Model}: {Teaching} {Reward} {Models} to {Know} {What} is {Unknown}}},
    shorttitle = {Uncertainty-aware {Reward} {Model}},
    url = {http://arxiv.org/abs/2410.00847},
    doi = {10.48550/arXiv.2410.00847},
    urldate = {2025-07-29},
    publisher = {arXiv},
    author = {Lou, Xingzhou and Yan, Dong and Shen, Wei and Yan, Yuzi and Xie, Jian and Zhang, Junge},
    month = feb,
    year = {2025},
    note = {arXiv:2410.00847 [cs]},
}

@misc{dorka2024Quantile,
    title = {Quantile {Regression} for {Distributional} {Reward} {Models} in {RLHF}},
    url = {http://arxiv.org/abs/2409.10164},
    doi = {10.48550/arXiv.2409.10164},
    urldate = {2025-07-29},
    publisher = {arXiv},
    author = {Dorka, Nicolai},
    month = sep,
    year = {2024},
    note = {arXiv:2409.10164 [cs]},
}

@misc{gulcehre_reinforced_2023,
    title = {Reinforced {Self}-{Training} ({ReST}) for {Language} {Modeling}},
    url = {http://arxiv.org/abs/2308.08998},
    doi = {10.48550/arXiv.2308.08998},
    urldate = {2026-01-28},
    publisher = {arXiv},
    author = {Gulcehre, Caglar and Paine, Tom Le and Srinivasan, Srivatsan and Konyushkova, Ksenia and Weerts, Lotte and Sharma, Abhishek and Siddhant, Aditya and Ahern, Alex and Wang, Miaosen and Gu, Chenjie and Macherey, Wolfgang and Doucet, Arnaud and Firat, Orhan and Freitas, Nando de},
    month = aug,
    year = {2023},
    note = {arXiv:2308.08998 [cs]},
}

@inproceedings{wei2022ChainThought,
author = {Wei, Jason and Wang, Xuezhi and Schuurmans, Dale and Bosma, Maarten and Ichter, Brian and Xia, Fei and Chi, Ed H. and Le, Quoc V. and Zhou, Denny},
title = {Chain-of-thought prompting elicits reasoning in large language models},
year = {2022},
isbn = {9781713871088},
publisher = {Curran Associates Inc.},
address = {Red Hook, NY, USA},
booktitle = {Proceedings of the 36th International Conference on Neural Information Processing Systems},
articleno = {1800},
numpages = {14},
location = {New Orleans, LA, USA},
series = {NIPS '22}
}

\appendix

\section{Reinforcement Learning for LLMs}
\label{subsection:methodology:ppo}


The goal in RL-optimization is to find the parameters $\theta$ that maximise the expected discounted return starting from the initial state under the policy $\pi_\theta$:
\begin{align}
\small
   J(\theta)=\E_{\pi_\theta}\sum_{t=1}^{T}\gamma^tr_t,
\end{align}

where $T$ signifies a point where the chosen token is the end-of-sequence token.
The policy gradient theorem~\cite{sutton1999policygradient} provides the formulation of gradient for $\pi_\theta$

\begin{align}
\small
    \nabla_\theta J(\theta) = \E_{\pi_\theta}{\sum_{t=0}^{T-1} \nabla_\theta \log \pi_\theta(a_t|s_t) A_{\pi_\theta}(s_t,a_t)}
    \label{equation:policy_gradient}
\end{align}
where  \(A_{\pi}\left(s, a\right)\) is the advantage function for policy $\pi$. 
Instead of directly working with the above gradient, PPO uses a gradient of the surrogate loss, which approximates the direction of the gradient above. 
The loss applies clipping and a KL penalty to prevent the small changes in the parameters $\theta$ from having unexpected changes on the policy $\pi_\theta$. 
For full details, please refer to~\cite{schulman2017ProximalPolicy} and Appendix~\ref{sec:appendix:ppo}.
  
\subsection{Bradley-Terry reward model}
\label{subsection:methodology:bt}

When the policy choses the end of sequence token the system receives the reward from the reward model $r_T=R_{\phi}(s_{T-1} \odot a_{T-1})$, where $s_{T-1}$ is the current context and $a_{T-1}$ is the end of sequence token.  $R_{\phi}(s)$ is another language model optimised with the Bradley-Terry~\citep{bradley1952RankAnalysis} loss
\begin{align}
    \small
    \begin{split}
        \Loss{h^1,h^2 ; \phi} = - \E_{\{\prompt,h^1,h^2\}} \log (
            & \sigma ( R_\phi(\prompt \odot h^1) \\
            & - R_\phi(\prompt \odot h^2))
        ),
    \end{split}
    \label{equation:bradley_terry}
\end{align}

where hypothesis $h^1$ is preferred over hypothesis $h^2$ for query \(\prompt\). Since the reward model is utilised only in the ultimate step, for its estimation, only the complete sequences are used. 

\subsection{Direct Preference Optimisation}
\label{subsection:methodology:dpo}

Given a preference dataset and assuming the Bradley-Terry preference model, the policy can be directly optimised with the \emph{direct preference optimisation}~(DPO) loss given by
\begin{align}
    \small
    \begin{split}
        \Loss{h^1, h^2; \theta} = & - \E_{\{\prompt,h^1,h^2\}} \Big[ \log \sigma \Big( \beta \log\frac{\pi_\theta(h^1|\prompt)}{\pi_{\mbox{ref}}(h^1|\prompt)} \\
        & - \beta \log\frac{\pi_\theta(h^2|\prompt)}{\pi_{\mbox{ref}}(h^2|\prompt)} \Big) \Big]
    \end{split}
    \label{equation:dpo_loss}
\end{align}

where for a query, \(\prompt\) hypothesis $h^1$ is preferred over hypothesis $h^2$, $\pi(h|\prompt) \approx \pi(h_T|\prompt \odot h_{0:T-1})$ and $\pi_{\mbox{ref}}$ is the reference LLM before DPO is applied. 
DPO bypasses the need to first learn the reward model and then perform reinforcement learning. 
Note, however, that discounting of the reward does not feature in DPO.

Direct preference optimisation bypasses the need for reinforcement learning by performing supervised learning with the objective given in Eq~\ref{equation:dpo_loss}. 
Practically, it is still beneficial to perform some label smoothing to have better-calibrated responses.

\subsection{Proximal policy optimisation}
\label{sec:appendix:ppo}

The advantage function $A_\pi(s,a)$ denotes the difference between the \(Q\)-function $Q_\pi(s,a)$, which is the expected return when taking action \(a\) in state \(s\) and from then on following the policy $\pi$ and the value function $V_\pi(s)$, which is the  expected return of state \(s\) for policy $\pi$:
\begin{align}
    \small
    \begin{split}
        A_\pi(s,a) &=Q_\pi(s,a)-V_\pi(s)\\
        Q_\pi(s,a) &=\E_\pi\sum^{T-1}_{t=0}\gamma^lr_{t+1}|s_t=s,a_t=a \\
        V_\pi(s) &=\E_\pi\sum^{T-t}_{l=0}\gamma^lr_t|s_t=s \\
    \end{split}
\end{align}

The surrogate loss, consisting of the clipping term and the KL-penalty term, is given by:
\begin{align}
    \small
    \begin{split}
        \mathcal{L}^{\text{CLIP+KL}}(\theta) & = \mathbb{E}_t \Big[
        \min \Big(
        \frac{\pi_\theta(a_t \mid s_t)}{\pi_{\theta_{\text{old}}}(a_t \mid s_t)} \hat{A}_t, \\ 
        & \text{clip}\left( \frac{\pi_\theta(a_t \mid s_t)}{\pi_{\theta_{\text{old}}}(a_t \mid s_t)}, 1 - \epsilon,\ 1 + \epsilon \right) \hat{A}_t
        \Big) \\
        & - \beta \cdot \sum_{a} \pi_{\theta_{\text{old}}}(a \mid s_t) \log \left( \frac{\pi_{\theta_{\text{old}}}(a \mid s_t)}{\pi_\theta(a \mid s_t)} \right)
        \Big],
    \end{split}
    \label{equation:surrogate_loss}
\end{align}
where $\text{clip}(x,a,b)$ ensures that $x$ is within the interval $[a,b]$.
The gradient of the surrogate loss (Eq~\ref{equation:surrogate_loss}) approximates the direction of the policy gradient~(Eq~\ref{equation:policy_gradient}). 

To estimate the advantage function $\hat{A}_t$ needed for the surrogate loss, generalised advantage estimation is used:

\begin{align}
    \small
    \begin{split}
    \hat{A}_t &= \sum_{l=0}^{T - t - 1} (\gamma \lambda)^l \, \delta_{t + l}
    \quad \text{where} \\
    \quad
    \delta_t &= r_t + \gamma V_\pi(s_{t+1}) - V_\pi(s_t)
    \end{split}
\end{align}

Parameter $\lambda \in [0,1]$ trades off bias (lower $\lambda$) and variance (higher $\lambda$). 
$V_\pi$ is initialised with the reward model $R_\phi$ for complete sequences, and parameters $\phi$ are updated in such a way so that $V_\pi$ estimates observed returns for partial sequences. 
In this way, the critic parameterised with $\phi$ influences the actor parameterised with $\theta$.

PPO is an online learning algorithm, which means it needs to adopt exploration during the process of optimisation. 
That is achieved by applying a temperature parameter to the softmax layer of policy $\pi_\theta$ to induce more varied responses.

\section{Training Configurations}
\label{sec:appendix:hyp}

Here, we give relevant configurations to reproduce the results presented in this paper.
The PPO configurations are given in \Cref{tab:hyp:ppo} and are used for every RLSF with PPO set-up. 
The DPO configurations are given in \Cref{tab:hyp:dpo} and are used in every RLSF with DPO configuration as well as for AfD. 
The supervised fine-tuning is used for AfD (Table~\ref{tab:hyp:sft}).
\begin{table}[h]
    \centering
    \begin{tabular}{ll}
        \toprule
        \textbf{Parameter} & \textbf{Value} \\
        \midrule
        Learning rate & 5e-5 \\
        Epochs & 5 \\
        Temperature for exploration & 0.7 \\
        KL coefficient $\beta$ & 0.05 \\
        PPO clipping range $\epsilon$ & 0.2 \\
        Discount factor $\gamma$ & 0.98 \\
        General advantage estimation $\lambda$ & 0.95 \\
        \bottomrule
    \end{tabular}
    \caption{PPO Training Hyperparameters}
    \label{tab:hyp:ppo}
\end{table}
\begin{table}[h]
    \centering
    \begin{tabular}{ll}
        \toprule
        \textbf{Parameter} & \textbf{Value} \\
        \midrule
        Learning rate & 5e-5 \\
        Epochs & 5 \\
        Label smoothing & 0.01 \\
        DPO $\beta$ & 0.2 \\
        \bottomrule
    \end{tabular}
    \caption{DPO Training Hyperparameters}
    \label{tab:hyp:dpo}
\end{table}
\begin{table}[h]
    \centering
    \begin{tabular}{ll}
        \toprule
        \textbf{Parameter} & \textbf{Value} \\
        \midrule
        Learning rate & 5e-5 \\
        Epochs & 5 \\
        \bottomrule
    \end{tabular}
    \caption{Supervised Fine-tuning Hyperparameters}
    \label{tab:hyp:sft}
\end{table}
%
\section{Impact of the Discount Factor $\gamma$}
\label{section:appendix:discount_factor}

See \Cref{table:discount_factor} for a comparison of different discount factors in PPO within our RLSF framework.

\begin{table}[h!]
    \centering
    \resizebox{0.95\linewidth}{!}{%
    \begin{tabular}{l cc cc}
        \toprule
        \textbf{Variant} 
        & \multicolumn{2}{c}{\textbf{MultiArith}} 
        & \multicolumn{2}{c}{\textbf{GSM8K}} 
        \\
        & Accuracy \(\uparrow\) & ECE \(\downarrow\) 
        & Accuracy \(\uparrow\) & ECE \(\downarrow\) 
        \\
        \midrule
        DPO & 96.13 & 10.52 & 84.74 & 17.43 \\
        \midrule
        PPO \((\gamma=1.0)\) & 98.52 & 8.12 & 87.13 & 12.49 \\
        PPO \((\gamma=0.98)\) & 98.83 & 7.81 & \textbf{88.14} & 12.54 \\
        \bottomrule
    \end{tabular}}
    \caption{
        Discount factor $\gamma$. Gemma 2 is fine-tuned using RLSF with PPO.
        \label{table:discount_factor}
    }
\end{table}

\end{document}